%% file: sample-sigconf.tex
\documentclass[sigconf]{acmart}

\usepackage{times}
\usepackage{soul}
\usepackage{url}
\usepackage[utf8]{inputenc}
\usepackage{graphicx}
\usepackage{xcolor}
\usepackage{dsfont}

\usepackage{amsmath,amsthm,amsfonts,amssymb}
\usepackage{booktabs}
\usepackage{algorithm}
\usepackage{caption}
\usepackage{subfigure}
\usepackage{enumitem}
\usepackage{wrapfig}
\usepackage{makecell}
\usepackage[noend]{algpseudocode}
\usepackage{multirow}
\usepackage{tabularx}
\usepackage{colortbl}
\usepackage{xcolor}
\usepackage{array}
\usepackage{color}
\usepackage{soul}

\newcommand*{\pr}{\textnormal{Pr}}

\newcommand*{\bt}{\boldsymbol{\theta}}

\newcommand*{\bx}{\mathbf{x}}

\usepackage{algorithmicx}  
\usepackage{algpseudocode}

\newtheorem{theorem}{Theorem}

\AtBeginDocument{%
  }

\setcopyright{acmcopyright}

\acmConference[XXX]{Make sure to enter the correct
  conference title from your rights confirmation emai}{XXX}{XXX}
\acmBooktitle{XXX} 
\acmPrice{XXX}
\acmISBN{XXX}

\begin{document}

\title{Interpretable and Effective Reinforcement Learning for Attacking against Graph-based Rumor Detection}






\author{
Yuefei Lyu
\and
Xiaoyu Yang\and
Jiaxin Liu\and
Philip S. Yu\and
Sihong Xie\and
Xi Zhang \\
}

\begin{abstract}
Social networks are frequently polluted by rumors, which can be detected by
advanced models such as graph neural networks.
However, the models are vulnerable to attacks and
understanding the vulnerabilities is critical to rumor detection in practice.
To discover subtle vulnerabilities,
we design a powerful attacking algorithm to \textit{camouflage} rumors in social networks based on reinforcement learning that can interact with and attack any black-box detectors.
The environment has exponentially large state spaces, high-order graph dependencies, and delayed noisy rewards, making the state-of-the-art end-to-end approaches difficult to learn features 
as large learning costs and expressive limitation of graph deep models.
Instead, we design domain-specific features to avoid learning features
and produce interpretable attack policies.
To further speed up policy optimization, we devise:
\textbf{(i)} a credit assignment method that decomposes delayed rewards to atomic attacking actions proportional to the their camouflage effects on target rumors;
\textbf{(ii)} a time-dependent control variate to reduce reward variance due to large graphs and many attacking steps,
supported by the reward variance analysis and a Bayesian analysis of the prediction distribution.
On three real world datasets of rumor detection tasks,
we demonstrate:
\textbf{(i)} the effectiveness of the learned attacking policy compared to rule-based attacks and current end-to-end approaches;
\textbf{(ii)} the usefulness of the proposed credit assignment strategy and variance reduction components;
\textbf{(iii)} the interpretability of the policy when generating strong attacks via the case study.
\end{abstract}

\ccsdesc[500]{Information systems~Data mining}
\ccsdesc[500]{Computing methodologies~Artificial intelligence}

\keywords{Graph adversarial attack, reinforcement learning, graph convolutional network, rumor detection}
\maketitle

\input{introduction}
\input{related}
\input{method}

\input{experiments}

\section{Conclusion}
In this paper, we propose AdRumor-RL, an interpretable and effective hierarchical attack framework against GCN-based rumor detector. We define a practical attack object with realistic constraints and use reinforcement learning to realize black-box attacks. Interpretable attacking features are designed to capture graph dependencies and ranking dependencies. To speed up learning, we design a credit assignment method to speed up learning and a time-dependent baseline to reduce variance. This attack framework can be extended to more applications in social networks.


\bibliographystyle{ACM-Reference-Format}
\bibliography{sample-base}

\appendix
\input{appendix}

\end{document}

%% file: introduction.tex
\section{Introduction}
Social networks, such as Twitter and Weibo, help propagate useful information.
However, they are also exploited to spread misinformation, such as rumors,
to manipulate opinions in a large scale. 
Detecting misinformation is important to trustworthy social networks.
Graph Convolutional Networks (GCN) \cite{12_DBLP:conf/iclr/KipfW17} can aggregate neighborhood information to deliver high detection accuracy~\cite{1_ijcai2020-197,3_Bian_Xiao_Xu_Zhao_Huang_Rong_Huang_2020,4_lu-li-2020-gcan}.
However, GCN is fragile to graph adversarial attacks.
For example, \cite{6_pmlr-v80-dai18b} showed that GCN are vulnerable to edges or features flipping that degrade node classification accuracy.
Focusing on rumor detection, \cite{1_ijcai2020-197}
restricts rumor producers to controlled accounts to camouflage rumors to be less suspicious through re-posting, following and commenting, to deceive a GCN detector.

To re-design more robust GCN-based rumor detectors,
it is critical to discover and understand the vulnerabilities
using an attacking model to simulate camouflage actions of various level of sophistication under realistic constraints.
However, existing attacking models, especially those based on deep neural networks and reinforcement learning, are too complicated to help humans understand how attacks are generated and what detector vulnerabilities are exploited.
We argue that the simplicity of the attacking models should be as important as their effectiveness to be useful to the detector designers.

\paragraph{Threat model}
To capture how rumors can be spread,
we use a heterogeneous graph consisting of nodes representing user accounts, messages, and comments, and edges representing posting, re-posting, and commenting.
An attacker controls some accounts to post messages and follow other accounts~\cite{7_NEURIPS2020_32bb90e8}.
All these operations can be represented as adding edges to the graph.
For each message node $v_i$, a trained GCN $f$ outputs $f(v_i)$ as the ranking of 
$v_i$ based on its suspiciousness,
and messages with high ranking will be removed as rumors.
During a time period,
only a few high influence messages can receive the most attention~\cite{Hodas2012}.
A high influence rumor is more useful for spreading misinformation, worth to be camouflaged,
as it has zero influence once detected.
Inspired by the ranking metric Normalized Discounted Cumulative Gain (NDCG),
we assume that an attacker aims to minimize the following objective function:
\begin{equation}
    \label{eq:ndcg}
    J = \frac{1}{Z} \sum_{i=1}^n\frac{w_i}{\log(f(v_i) + 1)} \mathds{1}[f(v_i)>m] ,
\end{equation}
where $Z$ normalizes the sum to $[0,1]$ and $n$ is the number of target rumors (not including other rumors that are not controlled by this attacker).
$w_i$ is the influence or weight of the $i$-th target rumor $v_i$ estimated in various ways~\cite{Bakshy2011}.
The indicator function $\mathds{1}[f(v_i)>m]$ truncates the contribution of target rumors whose suspiciousness ranking is greater than $m$.
Prior graph adversarial attacks assumed the target detectors are white-box~\cite{8_10.1145/3219819.3220078,9_12_DBLP:conf/iclr/ZugnerG19,10_ijcai2019-669} so that gradient-based attacks can be crafted.
In contrast, we assume that the architecture or model parameters of $f$ is unknown and only allow attackers having access to and mastering part of nodes in the social networks.

\begin{figure}
\centering
    \begin{minipage}[t]{0.32\linewidth}
        \centering
        \includegraphics[scale=0.49]{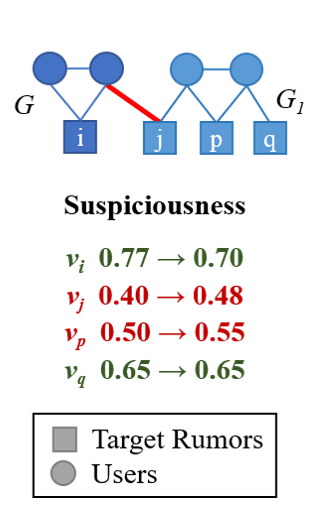}
        \text{\small (a) Graph dependencies}
    \end{minipage}
    \begin{minipage}[t]{0.66\linewidth}
        \centering
        \includegraphics[scale=0.49]{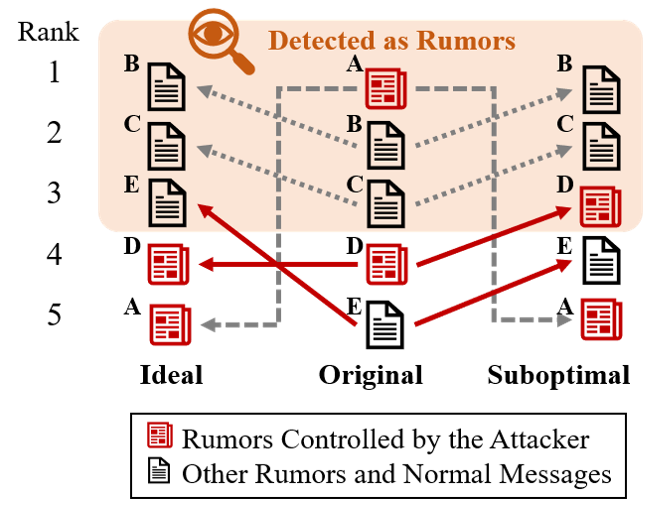}
        \text{\small (b) Ranking dependencies}
    \end{minipage}
    \caption{\small
    (a) Graph dependencies. When connecting $G$ to $G_1$, $v_i$ in $G$ becomes less suspicious and $v_j$ and $v_p$ in $G_1$ become more suspicious due to information propagation.
    The more remote $v_q$ is not affected.
    (b) Dependencies due to relative ranking of target rumors.
    \textit{Middle ranking}: top 3 messages are detected as rumors.
    \textit{Right ranking}: after an attack, rumor A escapes the detection while rumor D is detected.
    \textit{Left ranking}: after a better attack, rumors A and D both escape the detection.
    }
    \label{fig:dependencies}
\end{figure}

\paragraph{Challenges}
Reinforcement learning (RL) has been adopted
to learn from a sequence of attack actions and the black-box detector's output as rewards.
Prior RL-based attacks~\cite{6_pmlr-v80-dai18b} train neural networks as policies end-to-end.
Though effective, the neural RL policies are not interpretable~\cite{Ribeiro2016,Alharin2020} and need a number of iterations and trajectories for learning features.
Furthermore, as shown in Figures~\ref{fig:dependencies},
the dependencies among the messages on the graph and their relative suspiciousness ranking positions in objective (Eq. (\ref{eq:ndcg})) are exploitable vulnerabilities that are hard for the end-to-end approaches to learn from noisy feedbacks with large action spaces and long horizons.
The limited expressiveness of GCN with local feature aggregation~\cite{Garg2020, DBLP:conf/cikm/LeeRKKKR19, DBLP:conf/iclr/OonoS20} makes it difficult to learn global node influence and neighborhood topology that help effective camouflage.
Lastly,
training RL agent to operate on graph data is highly sample inefficient: taking an action at some state has weak correlation with future returns, when there are multiple attacking steps over large action spaces derived from graphs~\cite{wang2020long}; two similar actions under the same state can lead to significantly different returns since GCN is sensitive to slight perturbations~\cite{Ruiz2021GraphNN}. 
There are high reward variances over the large action space/horizon, and the prior state-based or action-based control variate fails to reduce the variance as the states/actions in a large graph can hardly be summarized by state-action vectors.

\paragraph{Proposed solution}
We propose AdRumor-RL to generate interpretable and effective evasion attacks to camouflage high influence rumors and deceive a GCN-based rumor detector.
We formulate an episodic Markov Decision Process (MDP) and a hierarchical RL algorithm to attack the detector.
An action adds an edge in two levels:
the agent first selects two graphs (possibly identical) and then two nodes from the selected graphs to add an edge.
A return (cumulative rewards) at the end of an episode represents how well the sequence of added edges reduces the objective Eq. (\ref{eq:ndcg}).
We have two inventions:

First, to learn a strong and interpretable attacking policy,
we use domain knowledge about rumor spread to design inherent but hard-to-learn features to train a linear policy. 
In particular, we include two kind of features that capture the dependencies due to the graphs and the ranking-based objective.
The graph makes rumors depend on each other so that linking two nodes $v_i$ and $v_j$ can make other connected target rumors more detectable and thus reduce attack effectiveness. 
This situation is shown in Figure~\ref{fig:dependencies} (a).
The relative suspiciousness ranking of rumors creates another type of dependencies among the rumors: pulling a target rumor down the suspicious list can push another target rumor into the top on the list, as shown in Figure~\ref{fig:dependencies} (b).
Both types of dependencies are hard for a data-driven learning agent to capture and 
we design features to capture such dependencies to train strong and interpretable attacking policies.

Second, we propose the time-dependent credit assignment and baseline to cope with learning difficulties due to the large action space/horizon and confused graph representation.
We decompose the returns of multiple added edges to each action to reduce ambiguity of the returns.
The decomposition preserves the returns and proportionally associate the due effects to individual actions to speed up learning.
We design a novel time-based control variate to treat with the high reward variance due to many steps of manipulations over large action spaces.
It "clusters" the rewards dependent on the time of trajectory step to reduce the reward variance and then leads to a smaller prediction variance,
supported by the sample reward variance analysis and a Bayesian analysis of prediction distribution.


%% file: related.tex
\section{Related Work}

\noindent\textbf{Rumor Detection with GCNs}.
Recently, many rumor detection methods make use of GCN to mine message propagation networks and social relationship networks. \cite{3_Bian_Xiao_Xu_Zhao_Huang_Rong_Huang_2020} proposed Bi-GCN to explore the bottom-down and bottom-up propagation modes of rumors. \cite{4_lu-li-2020-gcan} modeled the potential interaction graph of the retweeting user of the source tweet, and used GCN to process user correlation information.


\noindent\textbf{Adversarial Attack on GCNs}.
From different perspectives, adversarial attack can be divided into:
poisoning attack \cite{11_pmlr-v97-bojchevski19a} and evasion attack \cite{10_ijcai2019-669}; 
untargeted attack \cite{9_12_DBLP:conf/iclr/ZugnerG19} and targeted attack \cite{6_pmlr-v80-dai18b}; 
white-box and black-box \cite{6_pmlr-v80-dai18b,7_NEURIPS2020_32bb90e8}.
\cite{1_ijcai2020-197} constructed a user-tweet-comment graph and proposed a graph adversarial learning framework, which makes GCN learn malicious rumor camouflage behaviors in social networks.
We focus on evasion attacks against a black-box pre-trained rumor detector, targeting at a set of target rumors.

\noindent\textbf{Reward baseline of reinforcement learning}.
The reward baseline (control variate) can reduce variance of Monte Carlo estimation effectively~\cite{Sutton1998,Greensmith2004VarianceRT}. 
The basic method is to use the constant baseline \cite{kimura1998reinforcement,williams1992simple}, like with the difference between the reward and the average reward~\cite{Sutton1998,marbach2001simulation}.
A common method is to use state-value functions as state-dependent control variates \cite{sutton2000policy,konda2000actor}. 
Recently, some action-dependent methods are proposed \cite{Tucker2018TheMO,Wu2018VarianceRF}.
In this work, we propose a time-dependent control variate as reward baseline.

%% file: method.tex
\section{Preliminaries and Problem Definition}
\subsection{Rumor Detection on Social Networks}
We construct an undirected heterogeneous graph ${\mathcal G}$=$({\mathcal V}, {\mathcal E})$. 
The node set is ${\mathcal V}=\{v \mid v \in {\mathcal M} \cup {\mathcal U} \cup {\mathcal C}\}$, 
where ${\mathcal M}$, ${\mathcal U}$ and ${\mathcal C}$ are the sets of messages, users and comments, respectively. 
The edge set is ${\mathcal E}=\{(v_i,v_j)\mid v_i\in{\mathcal V},v_j\in{\mathcal V}\}$. 
There is a relation mapping function $\psi:{\mathcal E}\to{\mathcal L}$ and ${\mathcal L}=\{l_1, l_2, l_3\}$ is the set of three particular relation types:
user-message $l_1$, message-comment $l_2$, and user-user $l_3$.
$l_1$ indicates a user posts or re-posts a message, 
$l_2$ means that a comment is appended to a message, and $l_3$ means that a user is connected to the author of a message when the user re-posts the message, or when two users re-post the same message. 
There are many communities, which correspond to a set of connected components $\{G_1, G_2, ..., G_m \}$ in ${\mathcal G}$, where each connected component $G_i=({\mathcal V}_i,{\mathcal E}_i)$ is called a \textit{subgraph} in the sequel.

We take the R-GCN model \cite{14_10.1007/978-3-319-93417-4_38} as the rumor detector $f$. R-GCN exploits edge types and it shows better performance than GCN in rumor detection on the heterogeneous graphs. 
The propagation step from layer $k$ to $k+1$ is
\begin{equation}
    \label{eq:rgcn}
    h_i^{(k+1)}=\sigma\left(\sum_{l\in {\mathcal L}}\sum_{j\in \mathcal{N}_i^l}\frac{1}{\mid \mathcal{N}_i^l\mid}
    W_l^{(k)}h_j^{(k)}+W_0^{(k)}h_i^{(k)}\right),
\end{equation}
where $h_i^{(k)}$ is the hidden state of node $v_i$ in the $k$-th layer of the nerual network. $\sigma$ is ReLU function and $\mathcal{N}_i^l$ denotes the set of neighbor nodes of $v_i$ connected by relation type $l\in {\mathcal L}$. $W_l^{(k)}$ is the weight matrix for $k$-th layer and relation type $l$.  
The last layer uses the sigmoid function to output probability over each message $v_i\in{\mathcal M}$.
The probabilities are used to rank the messages, with ranking position $f(v_i)$.
We train the model by minimizing a cross-entropy loss on labeled messages.

\subsection{Influence Calculation}
To calculate node influence, we first calculate user influence using PageRank on ${\mathcal G}_{user}$ containing user nodes and user-user relations
\begin{equation}
    \label{eq:userinf}
    w_j=\textnormal{PageRank}({\mathcal G}_{user}, v_j), 
    v_j\in{\mathcal U},
\end{equation}
and the influence of the message node $v_i$ is calculated as
\begin{equation}
    \label{eq:inf}
    w_i=\max_{u\in \mathcal{N}_i^{1}} \textnormal{PageRank}({\mathcal G}_{user}, u) + \frac{\mid \mathcal{N}_i^{1}\mid-1}{z_1} + \frac{\mid \mathcal{N}_i^{2}\mid}{z_2}, 
    v_i\in{\mathcal M},
\end{equation}
where $\mathcal{N}_i^{1}$ and $\mathcal{N}_i^{2}$ are the user and comment neighbors of $v_i$.
$z_1$ and $z_2$ indicate the maximum number of re-posting and comments of a message respectively.
Intuitively, high influence user connection, multiple and comments make the message influential.

\subsection{MDP and Reinforcement Learning}
An MDP consists of a state space ${\mathcal S}$,
an action space ${\mathcal A}$, a reward function $r(S_t, A_t)$, a state transition probability distribution $\pr(S_{t+1}|S_t,A_t)$.
Since we focus on finite horizon, a discounting factor is not needed.
In our application, at any time $t$, a state $S_t$ is the graph ${\mathcal G}_t$ and an action $A_t$ is a pair of nodes $(v_i, v_j)$ or a pair of subgraphs.
The goal of the attacker is to train an attacking policy $\pi_{\bt}(A|S)$ that connects nodes in $T$ steps to minimize Eq. (\ref{eq:ndcg}).
Since $f(v_i)$ is a black-box, Eq. (\ref{eq:ndcg}) cannot be minimized via gradient-based approaches.
The trajectory is denoted by $(S_0, A_0, S_1, \dots, A_{T-1}, S_T)$.
From samples of $T$-step trajectories by interacting with the environment, reinforcement learning uses the reduction in the objective as a reward to learn a policy $\pi$ that maximizes the reduction 
\begin{equation}
    \label{eq:total_delta_ndcg}
    \Delta NDCG=J(0)-J(T)
\end{equation}
where $J(T)$ is the NDCG value at the end of step $T$ and $J(0)$ is the NDCG value before attack using Eq. (\ref{eq:ndcg}).
We will learn the action-value function $Q_{\pi}(S_t, A_t)$ so that
\begin{equation}
    Q^\ast(s, a) = \max_{\pi}\mathbb{E}[R_T|S_t=s, A_t=a],
\end{equation}
where $R_T$ is the random variable representing the reduction in Eq. (\ref{eq:ndcg}) at the end of the $T$ steps, starting from the $t$-th step with state $s$ and action $a$.
If $t<T$, $R_T$ is a delayed reward.
We approximate the $Q$ function using a linear function
$Q(s, a|\bt)=\mathbf{x}(s, a)^\top \bt$ with $\mathbf{x}(s, a)$ being a vector representing the $(s,a)$ pair.
We sample triples $(s, a, r)$ to train $Q$: 
\begin{equation}
\label{eq:q_objective}
    \min_{\bt} \sum_{(s, a, r)\sim\mathcal{D}} (Q(s, a|\bt) - r)^2.
\end{equation}
The loss function is similar to that in DQN~\cite{Mnih2015},
but we use the Monte Carlo estimation of the reward rather than bootstrapping using a $Q$ function.
Following the LinUCB algorithm \cite{15_10.1145/1772690.1772758}: given a sample $(s, a, r)$, the optimal $\bt$ is
\begin{equation}
    \label{eq:optimal_theta}
    \bt^\ast=(\mathbf{X}^\top \mathbf{X}+\mathbf{I})^{-1}\mathbf{X}^\top \boldsymbol{r}
\end{equation}
where $\mathbf{X}$ is the matrix of all sample $\mathbf{x}(s, a)$ vectors as rows.
$\mathbf{I}$ is the identity matrix of proper shape and $\boldsymbol{r}$ is a column vector with each element that represents the reward corresponding to a row in $\mathbf{X}$. 

As more samples are collected as the agent interacts with the environment,
the matrix $\mathbf{X}$, the inverse $(\mathbf{X}^\top \mathbf{X}+\mathbf{I})^{-1}$,
and the vector $\boldsymbol{r}$ can be updated incrementally and asynchronously. In the end of each episode $e$, we update the policy with state-action vector and rewards of $T$ steps as
\begin{equation}
\label{eq:theta_update}
\begin{aligned}
    &\mathbf{A}_{e}=\mathbf{A}_{e-1}+\sum\nolimits_{t=1}^{T}{\mathbf{x}_t\mathbf{x}_t^\top}\\
    &\boldsymbol{b}_{e}=\boldsymbol{b}_{e-1}+\sum\nolimits_{t=1}^{T}{r_t\mathbf{x}_t}\\
    &\bt_{e}={\mathbf{A}_{e}}^{-1}\mathbf{b}_{e},
\end{aligned}
\end{equation}
and it is initialized with $\mathbf{A}_0=\mathbf{I}_d$ and $\mathbf{b}_0=\mathbf{0}_{d*1}$, where $d$ is the feature dimension.

A policy with exploration can be derived from the $Q$ function.
At each step $t$ of episode $e$, the policy $\pi_{\bt}$ chooses the action $a_t$ from the action space ${\mathcal A}(t)$ at time $t$ as
\begin{equation}
\label{eq:linucb}
a_t = \mathop{\arg\max}\limits_{a\in {\mathcal A}(t)}\mathbf{x}(s_t, a)^\top\bt^\ast + 
\alpha \sqrt{\mathbf{x}(s_t, a)^\top {\mathbf{A}}_{e-1}^{-1} \mathbf{x}(s_t, a)},
\end{equation}
where $\alpha$ is a hyper-parameter to control the exploitation and exploration trade-off.

\section{Method}

\subsection{The Attack Framework}

\begin{figure}
\centering
\includegraphics[width=\linewidth]{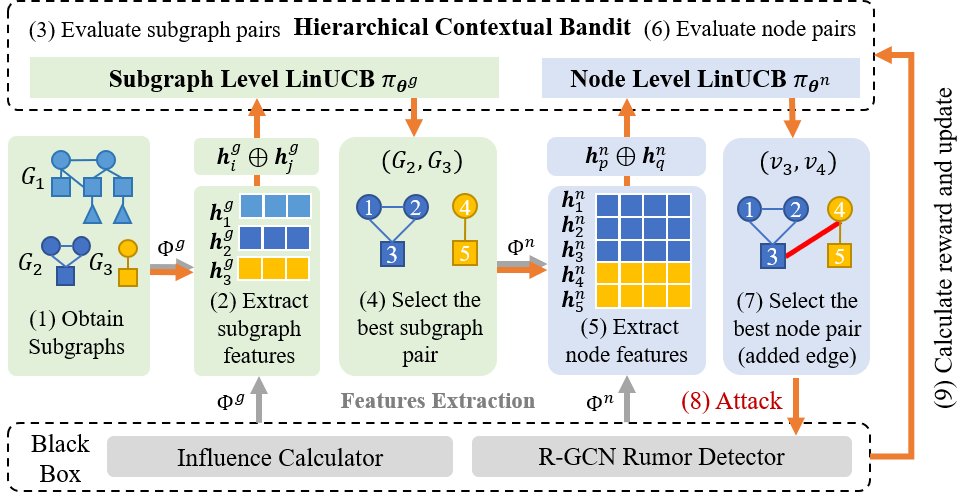}
\caption{\small The framework of AdRumor-RL. Following the orange arrows, there are 9 phases.
It extracts features of subgraphs and the features of two subgraphs are concatenated. Then the subgraph-level policy selects the best pair by Eq. (\ref{eq:linucb}). 
Phases 5-7 select a pair of nodes from the selected subgraphs for connection.
The selected edge is added for attack and the reward is calculated by Eq.~(\ref{eq:new_reward}), which is used to update parameters with Eq. (\ref{eq:theta_update}) after $T$ steps attack.
The grey arrows indicates the feature extraction process.
}
\label{fig:framework}
\end{figure}

The attacker can only have access to a subset ${\mathcal V}'\subset {\mathcal V}$ of controllable accounts and add edges in ${\mathcal E}' =\{(v_i,v_j) \mid v_i\in {\mathcal V}'\cap {\mathcal U}, v_j\in {\mathcal V}'\cap {\mathcal M}, (v_i,v_j)\notin {\mathcal E} \} $, i.e. connects controllable users to controllable messages. All rumors in ${\mathcal V}'$ constitute the target rumor set ${\mathcal O}$.

Due to the multi-communities characteristic of social networks, we design a hierarchical contextual bandit to decompose the adding edge action to subgraph level and node level. It is benefit to mine information of the user or message and its community, as well as reduce the action space to speed up reinforcement learning. The framework of AdRumor-RL we proposed is shown in Figure \ref{fig:framework}. 

On the subgraph level, we focus on the subgraphs $\{G_i=({\mathcal V}_i,{\mathcal E}_i)\}$ and extract their features with feature extraction method $\Phi^g$. The feature vector of $G_i$ is denoted by $\Phi^g(G_i)=\boldsymbol{h}^{g}_i=[h^{g}_{i,1}, h^{g}_{i,2},..., h^{g}_{i,d}]$, $d$ is the subgraph feature dimension. 
Two subgraphs are combined as an action, with the action space ${\mathcal A}_1=\{(G_i, G_j)\mid {\mathcal O}\cap {\mathcal V}_i\neq \emptyset, {\mathcal V}'\cap {\mathcal V}_j\neq \emptyset\}$, which means subgraph $G_i$ and $G_j$ must contain target rumors and controllable nodes respectively, to attack the target rumor using a controllable node.
Features of $G_i$ and $G_j$ are concatenated to
    $\mathbf{x}^{g}_{(i,j)}=
    \boldsymbol{h}^{g}_i\oplus\boldsymbol{h}^{g}_j
    =[h^{g}_{i,1},..., h^{g}_{i,d},h^{g}_{j,1},..., h^{g}_{j,d}]$
 and then the subgraph-level policy $\pi_{\bt^g}$ selects the best subgraph pair as shown in Eq. (\ref{eq:linucb}). 

On the node level, for each node in the selected subgraph pair $(G_i, G_j)$, we extract node features $\boldsymbol{h}^n$ with $\Phi^n$. Two nodes respectively from the selected two subgraphs are paired up to form the action space ${\mathcal A}_2=\{(v_p, v_q)\mid v_p\in {\mathcal V}_i, v_q\in {\mathcal V}_j, (v_p,v_q)\in {\mathcal E}'\}$. 
Similarly with the subgraph level, the node-level policy $\pi_{\bt^n}$ evaluates all node pairs with concatenated features $\mathbf{x}^{n}_{(p, q)}$ and decides the edge to be added for attack. 
With the attacked graph, we calculate the reward with Eq.~(\ref{eq:new_reward}) and update $\bt^g$ and $\bt^n$ with Eq.~(\ref{eq:theta_update}).

\subsection{Interpretable Attacking Feature}
\label{sec:feature}
Deep graph models, like GCNs, are usually used for feature extraction in graphs. They rely on localized first-order approximations of spectral graph convolutions 
and thus have difficulty in capturing high-order structures, such as node propagation influence and complicated structural patterns. 
Furthermore, some extra information, like message ranking and history attacking action, help learn effective camouflage policies.

Therefore, to capture effective information against rumor detector, we manually design interpretable features. 
These features help people understand the attacking policies and detector vulnerabilities.
It describes the social network on the subgraph and node level, and includes structural, social, influence, attack potential and ranking help message (RHM) features. 
Structural features are the basic characteristics of graph and node, like the number of nodes and edges, degree etc. 
Social features describe node type (rumor, non-rumor, user and comment) and ratios for different types of nodes. 
Influence features summarize user and message influence calculated in Eq. (\ref{eq:userinf}) and Eq. (\ref{eq:inf}).
Details are shown in Supplementary \ref{app.A}. 
Here we focus on attack potential and RHM features, associated with ranking and graph dependencies. We introduce two dependencies and describe corresponding designed features.

\subsubsection{Ranking dependencies}
\
\newline
\textit{Ranking dependencies} refers to the relativity of ranking. The drop ranking of one message leads to the risen ranking of other messages inevitably, which is called \textit{exchange} here.
For example, in the right column of Figure \ref{fig:dependencies} (b), attacking the target rumor A can lower its ranking, but also rises the ranking of another target rumor D. It is hoped to reduce the effects on the ranking of other targets when attacking a target rumor. 
Therefore, the non-target messages are expected to exchange with target rumors, 
such as the message E, because they don't affect the attack objective function in Eq. (\ref{eq:ndcg}). These non-target messages are named ranking help message (RHM). 

We design features for RHMs to capture ranking dependencies. RHM features are the classification probability of the RHMs in the selected subgraph or around the selected node before attack. When the probability of a RHM is similar to that of a target rumor, their rankings may also be close, so the RHM is more likely to exchange with the target rumor in ranking when they are connected. For example, if the ranking of the message E is 100 in  Figure \ref{fig:dependencies} (b), it is difficult for the message E to prevent the risen ranking of the target rumor D as shown in the left column. RHM features, i.e. classification probability, can help attackers identify whether a RHM has a chance to exchange with target rumors.

\subsubsection{Graph dependencies}
\
\newline 
\textit{Graph dependencies} occurs due to information propagations along the links in a graph. For example, when connecting the red edge in Figure \ref{fig:dependencies} (a), $v_i$ propagates the suspiciousness to $v_j$, which causes $v_i$ to be less suspicious while $v_j$ to be more suspicious. $v_i$ increases the attack performance and $v_j$ does the opposite. In addition, when $v_j$ is attacked directly, $v_p$ is also affected to be suspicious indirectly. Whether direct or indirect, it is expected to maximize positive effects and minimize negative effects. 

Therefore, we design the attack potential features to measure the effects when a target rumor is attacked:
i) Suspiciousness. We query the classification probability of target rumors in the selected subgraph or around the selected node before attack. 
Attacking suspicious rumors could change the NDCG more due to  small $f(v_i)$ in Eq. (\ref{eq:ndcg}). 
ii) Attack degree. We record the number of previous added edges within the subgraph and node neighbor. The object that has been attacked repeatedly will not produce much effects. 
iii) The number and distance of targets. It concerns the number of targets within node $k$-hop insides and their averaged distance to the selected node. It might have greater effects when attacking the target connected to more and closer other targets.
Furthermore, RHM features also play a role in graph dependencies because the RHM with low probability can propagate the credible information to the target rumor and lower the target rumor ranking when they are connected.

\subsection{Credit Assignment}
\label{sec:credit}

\begin{figure}
    \centering
    \subfigure{
        \includegraphics[width=0.98\linewidth]{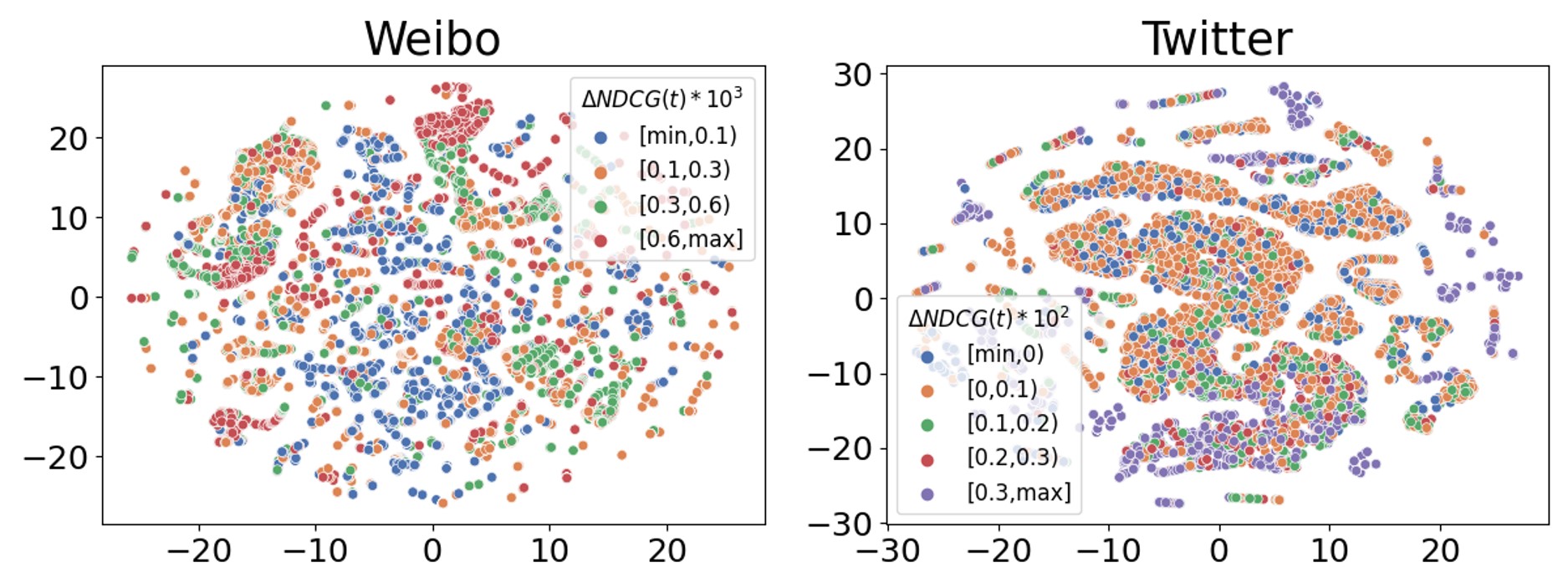}
    }
    \subfigure{
	    \includegraphics[width=0.98\linewidth]{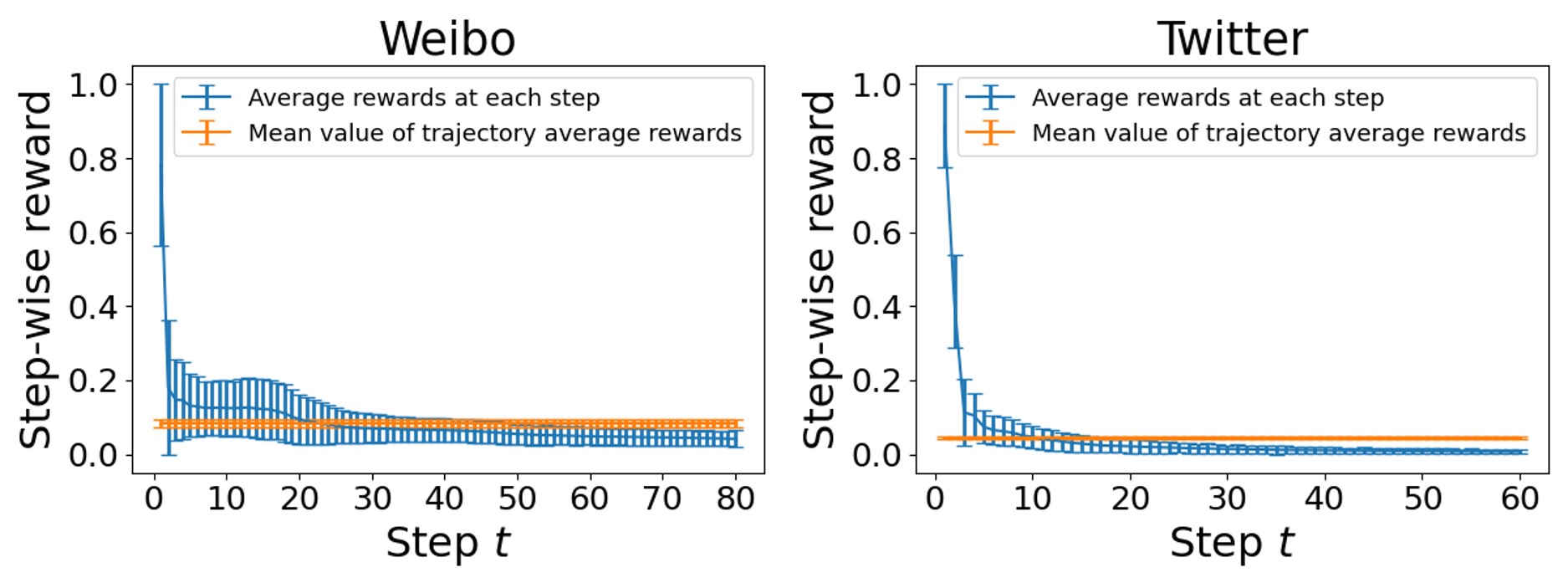}
    }
    
    \caption{\small The motivation of credit assignment and baseline design. \textit{Top}: The scatter plots of 2-D state-action vector processed by tSNE in Weibo and Twitter Dataset. 
    Scatters are colored according the value of $\Delta NDCG(t)$ (Eq. (\ref{eq:delta_ndcg})) in different intervals.
    Observe that vector representation of state-action cannot predict rewards well. 
    \textit{Bottom}: The average step-wise rewards at each step $t$ (Eq. (\ref{eq:baseline})) and the mean of average rewards of each trajectory ($\mathbb{E}_{\tau \sim {\pi_{\bt}}}\mathbb{E}_{t}[r_t]$). The former decreases as more edges are added. Vertical lines are standard deviations. The time-dependent baseline better predicts step-wise rewards and leads to more variance reduction. 
    }
    \label{fig:baseline_motivation}
\end{figure}
    

To learn to minimize the objective function in Eq. (\ref{eq:ndcg}), 
prior work \cite{credit1, credit2} shows that it is important to assign a proper reward as the feedback signal to individual action or state-action tuple that deserves the reward.
Otherwise, the policy will be trained to visit undesirable states or state-action tuples more frequently since the policy is unable to distinguish high and low-valued actions.
However, a state is a set of connected graphs and there can be exponentially many number of states, while a low-dimensional representation for distinguish any two states is extremely challenging~\cite{Garg2020,xu2018how}.
Furthermore, as shown in the top of Figure \ref{fig:baseline_motivation}, similar state-action vectors might corresponds to much different rewards. It means that the prior state-dependent or state-action-dependent rewards are not indicative of the state or action values. 

To address the above challenges,
we propose a time-dependent credit assignment method.
For a trajectory $(S_0, A_0, S_1, \dots, A_{T-1}, S_T)$, we define the following step-wise NDCG change:
\begin{equation}
    \label{eq:delta_ndcg}
    \Delta{NDCG}(t)=J(t-1)-J(t),t=1,2,...T,
\end{equation}
Due to telescoping, the overall reward after $T$ steps $\Delta{NDCG}$ (Eq. (\ref{eq:total_delta_ndcg})) is $\sum_{t=1}^{T}\Delta{NDCG}(t)$
and we can regard this equation as dividing the total reward $\Delta{NDCG}$ into individual rewards and assign them as credit to each step.

In the bottom of Figure \ref{fig:baseline_motivation}, we observe a significant difference in the step-wise rewards, especially between early and late steps.
Alternatively, if we assign the delayed return $R_T$ as a single reward to each $(S_t, A_t)$ of a trajectory for updating the policy as in Eq. (\ref{eq:q_objective}),
the values of late/early attacking steps are overestimated/underestimated, so that the agent won't learn to take high-value actions early on to maximize the overall return.
We assign $\Delta{NDCG}(t)$ to each step $t$ as
\begin{equation}
    \label{eq:reward}
    r_{t+1}=r(S_t,A_t)= o(\Delta{NDCG}(t+1)) ,
\end{equation}
where $o$ is Min-Max normalization function and the maximum and minimum value can be estimated by rule-based method described in section \ref{sec.exp_attack}. 
Compared with representing states of multiple graphs, the time has a simple representation, and the assigned credits are highly correlated with the time steps.

\subsection{Variance Reduction in Rewards}
\label{sec:critic}
\noindent\textbf{Reward baseline}.
Even with credit assignment, the action spaces at each step $\mathcal{A}_1(t)$ and $\mathcal{A}_2(t)$ are large and a policy is updated between trajectories,
the variance of the reward $r_t$ can be large.
To reduce the variance, is a common practice to subtract a baseline, or control variate, that highly correlates with the rewards~\cite{Weaver2001TheOR,Wu2018VarianceRF,Greensmith2004VarianceRT,Tucker2018TheMO,Mao2019VarianceRF}.
The new reward at step $t$ becomes
\begin{equation}
\tilde{r}_{t+1} = r_{t+1} - b(S_t, A_t),
\end{equation}
where $b(S_t, A_t)$ is the baseline that can depend on $S_t$~\cite{Greensmith2004VarianceRT}, $A_t$~\cite{Wu2018VarianceRF}, or some external input process~\cite{Mao2019VarianceRF}.
A simple baseline can be the average reward from a trajectory that is a constant:
\begin{equation}
\label{eq:const_baseline}
\overline{b} = \frac{1}{T}\sum_{t=1}^{T}r_t.
\end{equation}
The above constant baseline is not too much correlated with $r_t$ and thus too coarse to control the variance, as shown in the bottom of Figure~\ref{fig:baseline_motivation}.
Although a state-dependent or action-dependent baseline will correlate with $r_t$ better, the large state and action spaces and the lack of expressive representations of collection of graphs make such baselines hard to estimate.

\noindent\textbf{A time-dependent control variate}.
To address these difficulties, we propose the following time-dependent baseline 
\begin{equation}
    \label{eq:baseline}
    V^\pi (t) = \mathbb{E}_{\tau \sim {\pi_{\bt}}}[r_t],
\end{equation}
which is the expected rewards $r_t$ collected at step $t$ across trajectories and the expectation is taken over all possible state-action tuples that can appear at step $t$ when executing the policy $\pi_{\bt}$.
Figure~\ref{fig:baseline_motivation} (Bottom) plots $V^\pi (t)$.
We can see that the sample rewards $r_t$ decreases over time quickly, and compared with the constant baseline $\overline{b}$ in Eq. (\ref{eq:const_baseline}),
the time-dependent baseline $V^\pi (t)$ can predict $r_t$ more accurately over time.
As a result,
we use $V^\pi (t)$ as a control variate to reduce the variance in $r_t$,
leading to the following reward:
\begin{equation}
    \label{eq:new_reward}
    \tilde{r}_t = r_t - V^\pi (t).
\end{equation}
$\tilde{r}_t$ is used to replace $r_t$ to update policy in Eq. (\ref{eq:theta_update}).
$V^\pi(t)$ is defined for each $t$ across trajectories, rather than depending on rewards from other steps in the same trajectory.
The implicit assumption is that the rewards from different time steps are conditional independent.
Thus, we could analyze the reward variance with sample independence.


Compared with the constant baseline as Eq.~(\ref{eq:const_baseline}) and no baseline, the time-dependent baseline could reduce the variance in $\tilde{r}_t$ intuitively.
The time-dependent baseline could be seen as "clustering" the rewards according to the time $t$. 
It reduces the differences among reward clusters through subtracting the cluster center to shift the clusters to the close positions, and then reduces the reward variance.
\begin{theorem}
\label{the.var}
Given the sample matrix $\boldsymbol{R}=\{r_{e,t}\}\in\mathbb{R}^{E\times T}$, where $r_{e,t}$ indicates the element in the $e$-th row and the $t$-th column, 
and the matrix $\boldsymbol{R}'=\{r_{e,t}-b_t\}\in\mathbb{R}^{E\times T}$, 
where $b_t=1/E\sum_e{r_{e,t}}$ is the mean of the $t$-th column of $\boldsymbol{R}$. 
The variances of random variate in $\boldsymbol{R}$ and $\boldsymbol{R}'$ are denoted as $\sigma^2$ and ${\sigma'}^2$, 
we have $\sigma^2\geq{\sigma'}^2$.
\end{theorem}
The variance of random variate is denoted by $\textnormal{Var}(\cdot)$. Thus, we have $\textnormal{Var}(r_t-\overline{b})=\textnormal{Var}(r_t)\geq \textnormal{Var}(r_t-V^\pi (t))$.
The complete proof is shown in Supplementary \ref{app.Variance_analysis}.
In fact, the state-dependent or action-dependent baseline could also been seen as rewards "clustering" according to the state or the action.
However, they deviate the cluster center when learning the baseline due to the large state and action spaces and graph representation difficulties. 
By contrast, the time-based control variate is an unbiased estimation of the cluster center, so we regard the time as the "clustering" factor and apply the time-dependent baseline.

\noindent\textbf{Reduced variance in the predicted rewards}.
The linear function $\mathbf{x}^\top \bt$ is used to predict the future return starting state-action pair$(s,a)$ by following the policy parameterized by $\bt$.
The target variate is $\tilde{r}=\mathbf{x}^\top \bt+\epsilon$ and $\epsilon$ is the noise on the data. 
Assuming that $\epsilon$ is zero mean Gaussian noise with the precision (inverse variance) $\beta$,
we show that the reduced variance in the rewards can be transferred to reduced variance in future predictions. 
Using Bayesian ridge regression,
according to Eq. (3.57) and Eq.~(3.58) of \cite{prml},
the predictive distribution $p(r|\mathbf{x},\boldsymbol{r},\boldsymbol{S},\beta)$ at a new input $\bx$
has variance 
\begin{gather}
    \label{eq:pd_variance}
    {\eta}^2(\mathbf{x})=\frac{1}{\beta}+\mathbf{x}^{\top}\boldsymbol{S}\mathbf{x}
\end{gather}
where $\boldsymbol{S}$ is the variance of the posterior distribution of $N$ observed data. The posterior distribution becomes narrower when new data are observed so the second term goes to zero as $N\rightarrow\infty$ \cite{Qazaz1997}. A lower $1/\beta$ therefore leads to a smaller
linear regression variance.
Using maximum likelihood estimation, we can estimate the $\beta$ as in Eq. (3.21) of \cite{prml}:
\begin{equation}
    \label{eq:beta}
    \frac{1}{{\beta}_{ML}}=\frac{1}{N}\sum^N_{n=1}{(\tilde{r}_n-\bt^{\top}\mathbf{x})}^2= \textnormal{Var}(\tilde{r_t})
\end{equation}
since $\mathbb{E}[\tilde{r}_t]=\bt^\top \bx$. 
A reduced $\textnormal{Var}(\tilde{r}_t)$ directly leads to a reduced variance in the predicted reward in Eq. (\ref{eq:pd_variance}).

%% file: experiments.tex
\section{Experiments}
\label{sec:experiments}
\subsection{Datasets}
We conduct experiments on three real-world datasets: Weibo \cite{weibo}, Twitter \cite{TwitterDataset} and Pheme \cite{pheme}. They contain rumors and non-rumors, as well as user, retweeting and comment information. 
We randomly sample rumors in Pheme to construct a dataset with sample ratio 1:5, denoted by Pheme-, to simulate the realistic imbalanced scenarios. 
We split the datasets for rumor detector training and testing using a ratio of 7:3. 
The R-GCN detector we used is more effective than GCN as shown in Supplementary \ref{sec:exp_detector}. 
Reinforcement learning is to learn from experience, so we perform attacks and focus on results in training set. 
We randomly select 20\% of the authors and their messages in the training set as controllable nodes. There are 213, 171, 180 and 84 target rumors in Weibo, Twitter, Pheme, and Pheme-, respectively. Table \ref{tab:dataset} shows the dataset statistics. We can see that the graph consists of many subgraphs, which indicates many communities in social networks.

\begin{table*}
\begin{minipage}{0.35\linewidth}
\centering
    \small
    \caption{Statistics of dataset. 
    \textit{Subgraphs} are connected components.
    \textit{Authors} are the users who post messages.
    \textit{Retweeters} are the users who only retweet without posting messages.
    }
    \label{tab:dataset}
    \begin{tabular}{lrrrr}
    \toprule
        Dataset & Weibo & Twitter & Pheme & Pheme- \\ \midrule
        Nodes & 10280 & 3049 & 11950 & 9809  \\
        Edges & 16412 & 7206 & 14737 & 11678 \\ 
        Subgrphs & 2392 & 467 & 2450  & 2234\\  \midrule
        Rumors & 1538 & 981 & 1972  & 793 \\ 
        Non-rumors & 1849 & 1158 & 3830 & 3830  \\ 
        Authors & 2440 & 992 & 2837 & 2488 \\ 
        Retweeters & 4415 & 82 & 1496 & 1208 \\ 
        Comments & 38 & 0 & 1815 & 1490 \\ 
    \bottomrule
    \end{tabular}
    \end{minipage}
    \begin{minipage}{0.63\linewidth}
    \centering
    \small
    \caption{The attack performance results in the metric $\Delta NDCG (\times10^{-2})$. \textbf{boldfaced font} and $\ast$ mean the best performance and the runner-up among all methods respectively. }
    \label{tab:attack}
    \begin{tabular}{lrrrrrrrr}
    \toprule
        ~ & \multicolumn{3}{c}{Weibo} & \multicolumn{2}{c}{Twitter} & \multicolumn{2}{c}{Pheme} & Pheme- \\ \midrule
        Original & \multicolumn{3}{c}{58.70} & \multicolumn{2}{c}{63.49} & \multicolumn{2}{c}{45.14} & 31.83\\ \midrule
        Horizon & 20 & 40 & 80 & 30 & 60 & 30 & 60 & 30 \\ \midrule
        Random & 0.41 & 0.82 & 1.62 & 1.00 & 2.32 & 0.40 & 1.08 & 0.51  \\ 
        Random+ & 0.73 & 1.37 & 2.79 & 3.15 & 6.54 & 2.03 & 3.88 & 2.64  \\ 
        Degree & 0.44 & 0.88 & 1.87 & 14.75 & 18.58 & 2.17 & 4.84 & 5.63  \\ 
        Influence & 1.21 & 1.74 & 2.20 & $\ast$19.23 & $\ast$19.35 & 6.01 & 7.32 & 6.51  \\ 
        DCG & $\ast$2.30 & $\ast$3.14 & $\ast$3.98 & 18.99 & 19.34 & $\ast$6.13 & $\ast$7.38 & $\ast$6.56  \\ 
        GC-RWCS & 1.17 & 1.44 & 1.90 & 2.74 & 10.63 & 2.40 & 2.63 & 2.80  \\ 
        RL-S2V & 0.67 & 1.30 & 2.34 & 1.26 & 2.38 & 1.94 & 3.28 & 2.60  \\ 
        AdRumor-RL & \textbf{2.51} & \textbf{3.82} & \textbf{5.48} & \textbf{25.17} & \textbf{27.30} & \textbf{9.73} & \textbf{10.79} & \textbf{6.99}  \\  \bottomrule
    \end{tabular}
    \end{minipage}
\end{table*}
\subsection{The Performance of Attack}
\label{sec.exp_attack}
We measure the attack performance with $\Delta NDCG$ in Eq. (\ref{eq:total_delta_ndcg}) and compare our method to the following four rule-based attacking strategies and two state-of-the-art black-box attack methods:

\begin{itemize}[leftmargin=*]
\item
\noindent\textbf{Random and Random+}. It connects edges between users and messages randomly, denoted by Random. Inspired by \cite{1_ijcai2020-197}, we propose two variants GU-R and BU-N, denoted by Random+. GU-R connects edges between good users and target rumors randomly and BU-N connects edges between bad users (who post target rumors) and non-rumors randomly. The main idea is that adding edges as above helps the attacker camouflage rumors.

\item\textbf{Degree}. It selects the target rumor with the highest degree. There are also two variants GU-R and BU-N. The former connects the selected rumor with a random good user, and the latter connects the author of the selected rumor with a random non-rumor. High degree might usually mean high influence. This method expects to attack high influence rumor because the target rumor with large $w_i$ would sharply change the NDCG value in Eq. (\ref{eq:ndcg}).

\item\textbf{Influence}. It selects the target rumor with the highest influence value as Eq.~(\ref{eq:inf}). GU-R and BU-N are also two variants and work similarly with Degree.

\item\textbf{DCG}. It calculates rumor DCG value $w_i / \log(f(v_i)+1)$ of target rumor $v_i$ and selects the rumor with the highest DCG value. GU-R and BU-N are variants. DCG is closely related to the objective function in Eq. (\ref{eq:ndcg}) and is a strong baseline.

\item\textbf{GC-RWCS}\cite{6_pmlr-v80-dai18b}. It proposes a node selection strategy with a greedy procedure to calculate the importance score. It meets the black-box setting as well as limited access and attack. We use it to select candidate target rumors and apply GU-R and BU-N variants.

\item\textbf{RL-S2V}\cite{7_NEURIPS2020_32bb90e8}. It is a RL-based graph adversarial attack method. It represents the nodes with structure2Vec and makes use of two DQN to choose two nodes respectively within 2-hops of the target node, and then add or delete the edge between two nodes. To compare with our method, we modified the targeted attack to untargeted attack. Similar with DCG method, we calculate rumor DCG value and select the top-$T$ rumor with the higher DCG value as the target node for RL-S2V, and then conduct $T$-times attack. 
\end{itemize}
We average the results of 30 experiments for the rule-based methods. For AdRumor-RL, we use 10 random seeds for initialization. There are 1000 episodes in total with $\alpha$=1.0. We average the results of the last 100 episodes which tend to be stable. 
Results are shown in Table \ref{tab:attack} and we show the better one for the variants GU-R and BU-N. 
We can see that, 
i) AdRumor-RL has the best performance in all situations. 
ii) Some rule-based methods are strong baselines because they are superior to classical graph attacking models, especially DCG. For imbalanced Pheme- and $T$=20 in Weibo, the performance of some rule-based methods is close to AdRumor-RL. It means that they can identify the top attack targets effectively. 
iii) As $T$ increases, the improved performance of AdRumor-RL becomes more significant. AdRumor-RL is still effective even with large action space and long horizon.

\subsection{Effectiveness of Feature Design}
\begin{figure*}
    \centering
    \subfigure{
        \includegraphics[scale=0.19]{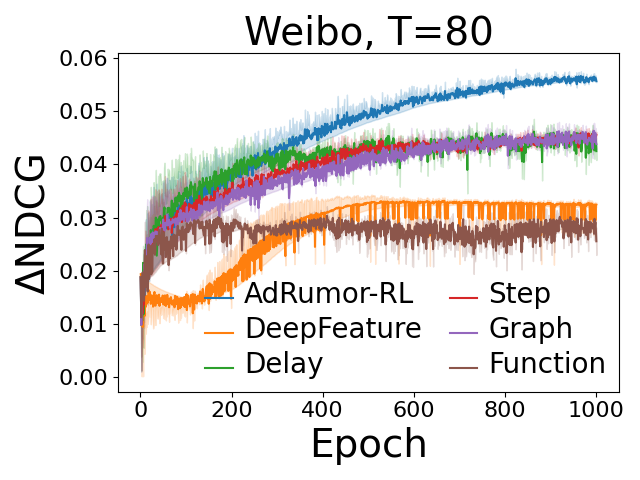}
    }
    \subfigure{
	    \includegraphics[scale=0.19]{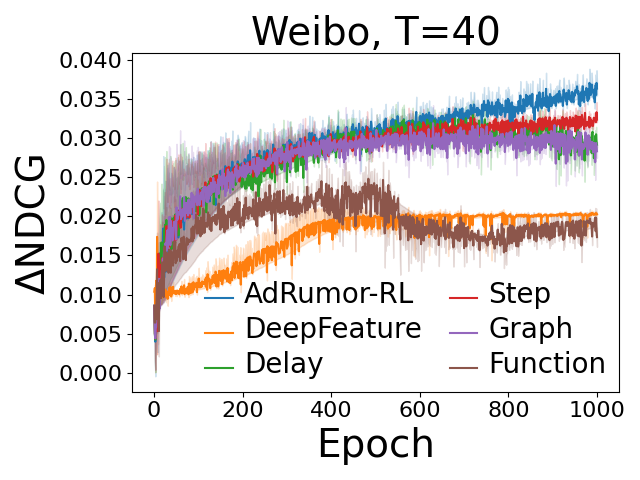}
    }
     \subfigure{
        \includegraphics[scale=0.19]{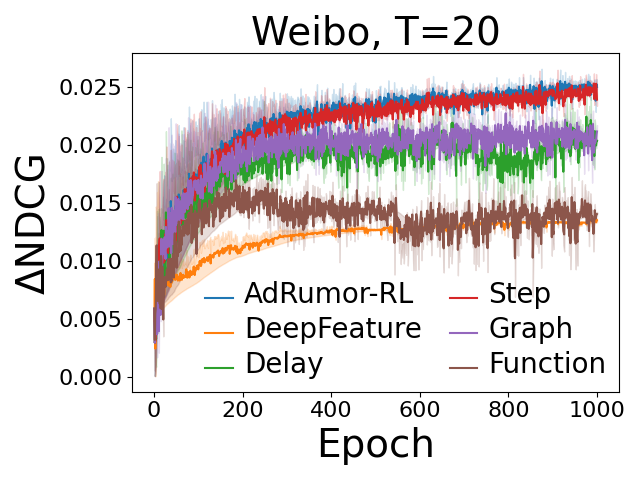}
    }
    \subfigure{
	    \includegraphics[scale=0.19]{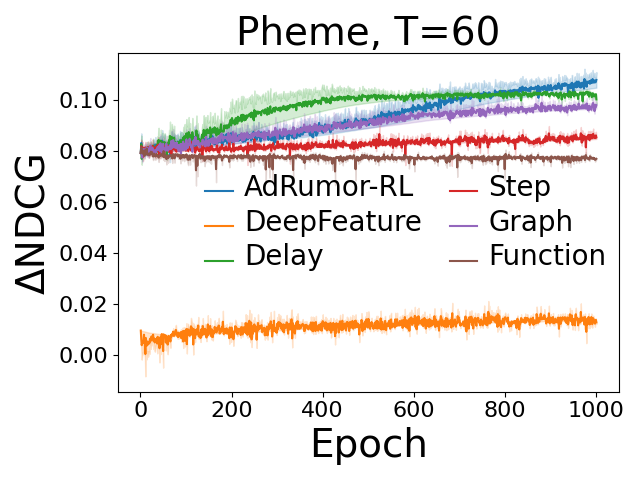}
    }
     \subfigure{
        \includegraphics[scale=0.19]{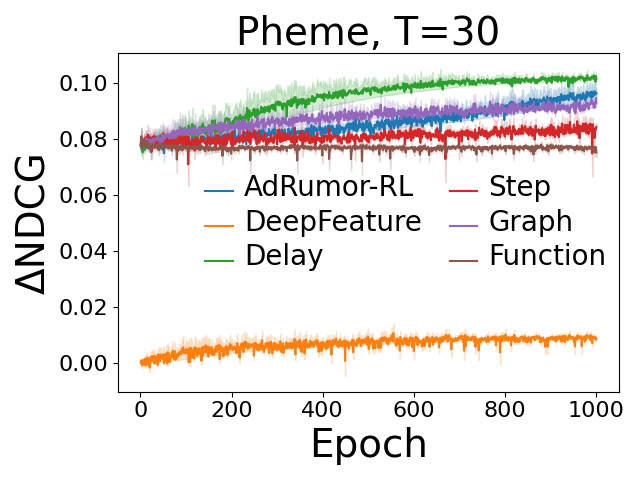}
    }

    \caption{Comparison experiments. Curves are smoothed and shadows show the standard variances.
    \textit{AdRumor-RL} is our method. 
    \textit{DeepFeature} extracts deep features. 
    \textit{Delay} regards the delayed return as reward. 
    \textit{Step} uses the mean value of all rewards as the constant baseline.
    \textit{Graph} uses the reward baseline that depends on history attack degree.
    \textit{Function} uses the state-dependent reward baseline.
    }
    \label{fig.line_compare}
\end{figure*}

\begin{figure}
    \centering
    \subfigure[Weibo, $T$=80]{
        \includegraphics[scale=0.25]{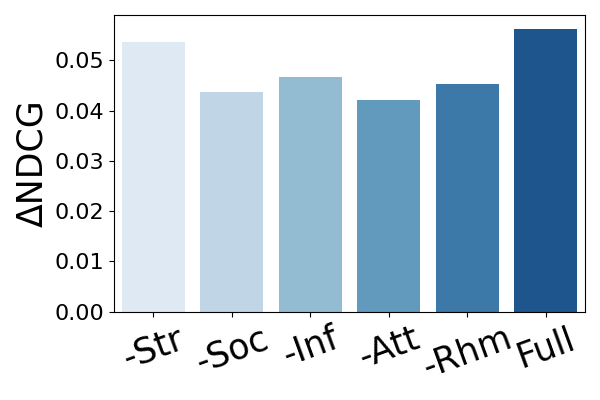}
    }
    \subfigure[Pheme, $T$=60]{
	    \includegraphics[scale=0.25]{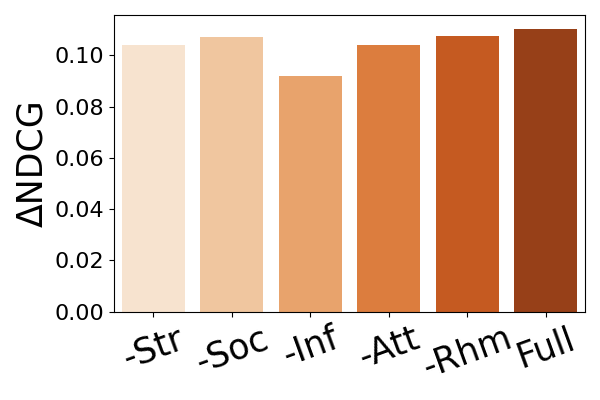}
    }
    \caption{Feature ablation experiments in Weibo ($T$=80) and Pheme ($T$=60). Elimination of structural, social, influence, attack potential and RHM features are named  \textit{-Str},  \textit{-Soc},  \textit{-Inf},  \textit{-Att} and  \textit{-Rhm}.  \textit{Full} indicates no elimination.
    }
    \label{fig.abla}
\end{figure}
\begin{figure}
    \centering
    \subfigure[Weibo, $T$=80]{
        \includegraphics[scale=0.25]{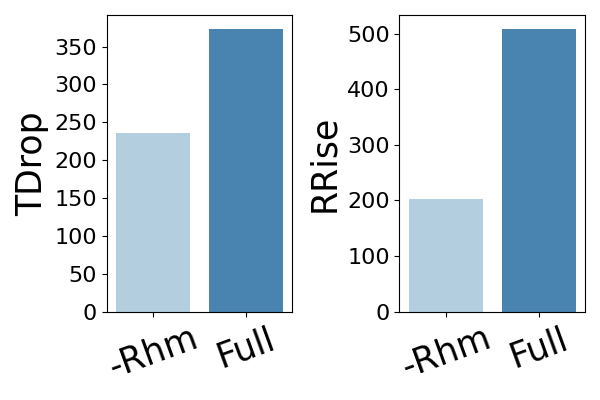}
    }
    \subfigure[Pheme, $T$=60]{
	    \includegraphics[scale=0.25]{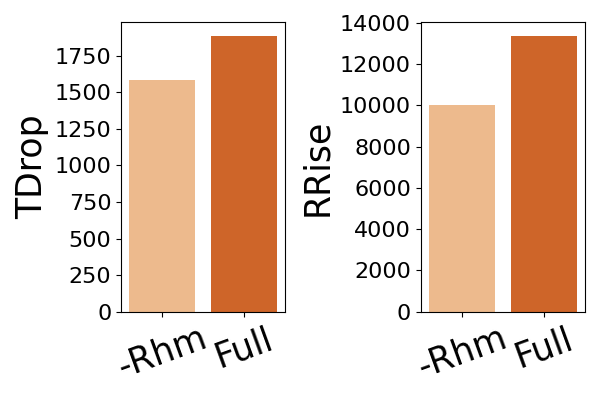}
    }
    \caption{The average TDrop and RRise values for all steps in best 100 episodes of each experiment in Weibo ($T$=80) and Pheme ($T$=60).
    }
    \label{fig.exp_ranking}
\end{figure}

\noindent\textbf{Superior to deep features}. 
We verify our designed features are better than end-to-end deep feature extraction. 
In phrases 2 and 5 of Figure \ref{fig:framework},
we replace our manually designed feature extraction method $\Phi$ with another R-GCN as features extraction model. In details, we use a R-GCN model similar with the detector, which outputs 64-dim node embedding in the last convolution layer, followed by a linear classification layer. It is pre-trained using labeled messages and is fixed during AdRumor-RL. It outputs node embedding as node features and averages node embeddings as graph features. The curve Deepfeature in Figure \ref{fig.line_compare} shows the results and it is far less effective than other methods with designed features.

\noindent\textbf{Feature ablation experiments}. 
We show the effectiveness of different type of features features as shown in Figure \ref{fig.abla}. 

\noindent\textbf{Graph and ranking dependencies.} 
We take the RHM features as an example to demonstrate its role on two dependencies.
RHMs help lower the ranking of target rumors or exchange with target rumors in ranking, which corresponds to graph and ranking dependencies, respectively. We show the effects with two metrics. 
\begin{itemize}[leftmargin=*]
\item Target rumor ranking drops (TDrop): the total drop in ranking positions of target rumors in the selected subgraphs of each step. 

\item Ranking help message rises (RRise): the total risen ranking positions of RHMs in the selected subgraphs of each step and these RHMs must reduce the target rumor rises. \end{itemize}
TDrop and RRise reflect the role of RHMs on graph and ranking dependencies respectively. In Figure \ref{fig.exp_ranking}, we can see higher TDrop and RRise value with using RHM features designed in Section \ref{sec:feature}.

\subsection{Effectiveness of Credit Assignment and Baseline Design}
\label{sec:exp_baseline}
\textbf{Credit assignment}. 
We use delayed return for comparison. It calculates $\Delta NDCG$ in Eq. (\ref{eq:total_delta_ndcg}) and normalizes it as each step reward. The curve Delay in Figure \ref{fig.line_compare} shows the results. Our method performs the best in most situations. As there are no much significant difference between the early and late steps with the short horizon, it is less necessary to focus on the step-wise effects, so Delay might perform similar to or better than our method due to good evaluation of episode overall effects.

\noindent\textbf{Reward baseline}. 
To show the effectiveness of the baseline we designed in Section \ref{sec:critic}, we compare our work with three baselines and results are shown in Figure \ref{fig.line_compare}.
\begin{itemize}[leftmargin=*]
\item\textbf{Step} is the common constant baseline \cite{Sutton1998,Weaver2001TheOR} as Eq.~(\ref{eq:const_baseline}). It records all history $r(S_t, A_t)$ and averages them as the baseline.

\item\textbf{Graph} is the baseline that depends on the number of previous added edges in selected subgraphs or ego networks of selected nodes, i.e. regarding them as "clustering" factor. 
It averages corresponding history $r(S_t, A_t)$ as the reward baseline. 
We find that attack degree is an important feature in experiments and design the baseline. 
It considers attack performance attenuation with manipulating the same objects.

\item\textbf{Function} is the common state-dependent method \cite{sutton2000policy,konda2000actor}, which learns the linear state-value function $V(\mathbf{x})=W'\mathbf{x}+b$ to predict the baseline, where $\mathbf{x}$ is the subgraph/node pair features. It updates parameters $W'$ and $b$ by minimizing the mean square error loss with $r(S_t, A_t)$ at each episode. Two functions are used on subgraph and node level.
\end{itemize}
As $T$ grows, the gap between AdRumor-RL and other baseline method becomes larger. Our time-based baseline effectively mitigates the negative effects of the long horizon.

\subsection{Case Study}
\begin{table}
    \centering
    \small
    \caption{Case study for feature importance. It lists the top-8 subgraph/node level feature according to the absolute value of LinUCB weight in the last episode of Weibo experiment that performs well. The subgraph/node pair is denoted by $(G_i,G_j)$/$(v_p,v_q)$. The features are shown in Supplementary \ref{app.A}.}
    \label{tab.feature_importance}
    \resizebox{\linewidth}{!}{
    \begin{tabular}{lrlr}
    \toprule
         Feature & Weight  & Feature & Weight  \\ \midrule
        \multicolumn{2}{c}{Subgraph Level}  &  \multicolumn{2}{c}{Node Level} \\\midrule
         1.$G_i$ n\_nodes                & 0.087    & 9.$v_p$ ego\_review\_ratio       & -0.057\\
        2.$G_j$ n\_edges                & -0.074    & 10.$v_q$ avg\_node\_attack\_degree & -0.024\\
         3.$G_j$ avg\_rhm\_suspicious     & -0.070    & 11.$v_q$ avg\_rhm\_suspicious     & -0.021\\
         4.$G_j$ non-rumor inf max      & 0.064    & 12.$v_p$ min\_neighbor\_suspicious & 0.019\\
         5.$G_i$ avg\_rhm\_suspicious     & 0.050   & 13.$v_p$ max\_rhm\_suspicious     & -0.017 \\
         6.$G_i$ review\_ratio           & -0.049    & 14.$v_p$ avg\_node\_attack\_degree & -0.016\\
         7.$G_j$ avg\_target\_suspicious          & -0.048    & 15.$v_q$ max\_rhm\_suspicious     & 0.015\\
         8.$G_i$ avg\_target\_suspicious          & 0.042   & 16.$v_q$ avg\_node\_suspicious     & -0.015\\
        \bottomrule
    \end{tabular}}
\end{table}
\noindent\textbf{Interpreting the feature importance}. 
Because we use linear function $\bt^\top\textbf{x}$ in LinUCB, we could correspond the each item of $\bt$ to one feature, and then evaluate how important the features are for decision making in an experiment episode. For example, Table \ref{tab.feature_importance} shows important subgraph and node level feature according to the feature weights. 
We could analyze these important features as following: 
for subgraph pair $(G_i,G_j)$ and node pair $(v_p,v_q)$, we could see that 
i) the size of subgraph plays an important role (index 1 and 2). 
ii) The suspicious RHM nodes in $G_j$ have a negative effect because they would propagate suspicious information to target rumors in $G_i$ (index 3 and 11). 
iii) The more suspicious target rumors in $G_i$ and less suspicious target rumors in $G_j$ contribute to better attack performance (index 7, 8 and 16). 
iv) The high attack degree of $v_p$ and $v_q$ is not conducive to attack (index 10 and 14). 

\begin{figure}
    \centering
    \includegraphics[width=0.95\linewidth]{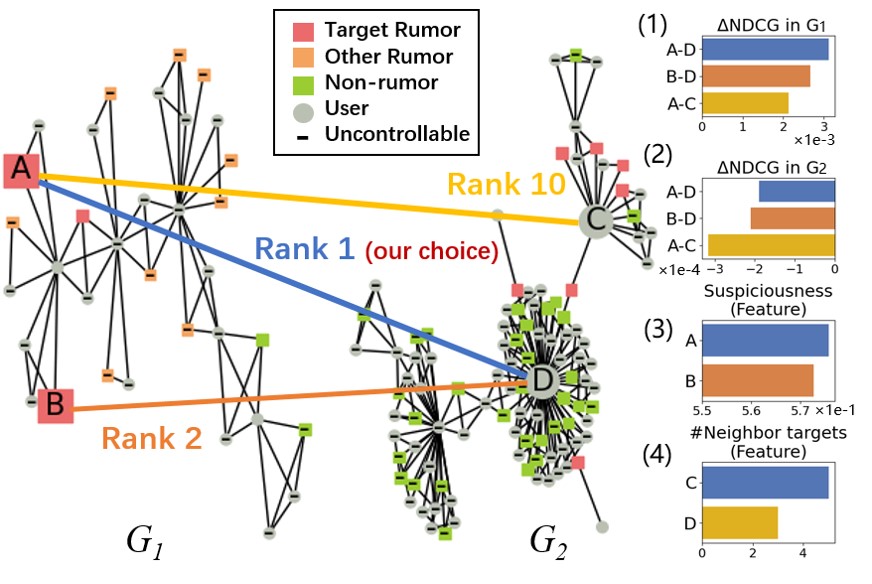}
    \caption{Case study for graph dependencies. \textit{Left}: The selected subgraph pair $(G_1,G_2)$ in Weibo and three edges with their performance ranking: A-D (rank 1), B-D (rank 2) and A-C (rank 10). \textit{Right}: (1) $\Delta NDCG$ of targets in $G_1$ (see target rumors in $G_1$ as targets in Eq. (\ref{eq:ndcg})). (2)$\Delta NDCG$ of targets in $G_2$. (3) The rumor probability of target rumor A and B. (4) The number of neighbor target rumors for the user C and D.
    }
    \label{fig.case}
\end{figure}

\noindent\textbf{Interpreting the feature effectiveness for graph dependencies}. 
We shows that AdRumor-RL captures the graph dependencies with designed features. In Figure \ref{fig.case}, we pick the subgraph pair $(G_1, G_2)$ of a step that performs well in the experiment and traverse all node level actions for $(G_1, G_2)$, and then display 3 edges and their $\Delta NDCG$ ranking. We can see that AdRumor-RL chooses the best pair A-D.
Suspiciousness drives the agent to choose the target rumor A in $G_1$ instead of B and it achieves the highest positive effects. The number of neighbor targets helps the agent choose the user D in $G_2$ instead of C, which reduces the lowest negative effects the most. These features help balance positive and negative effects under the situation of graph dependencies.

%% file: appendix.tex
\section{Detector Effectiveness}
\label{sec:exp_detector}
We attack against a R-GCN rumor detector with three hidden layers,
and compare it with a two-layer GCN model. The hidden dimension is 64 and the learning rate is 0.01.  
The performance is shown in Table \ref{tab:detector}. We could see that R-GCN improves the detection performance effectively.
We also calculate the global NDCG in training set, which see all training rumors as target rumors in Eq. (\ref{eq:ndcg}) and show the overall system performance. It shows that R-GCN could detect the high influence rumor better. 
It is worth mentioning that we train models with only graph structure for eliminating interference of text contents and other features. 

\section{Feature Design}
\input{appendix_table}
\label{app.A}
We design the subgraph and node features as shown in Table~\ref{tab.features}. 
On the subgraph, we describe the entire graph. On the node level, we focus on the node, its $k$-hop neighbor insides and the ego network. For the feature whose range is not $[0,1]$, we use Max-Min normalization. The hyper-parameter $k$ is 3 in our experiments. 
Through averaging the weights of multi experiments, we could observe the feature importance from the perspective of the entire dataset.

\begin{table}
    \centering
    \small
    \caption{The performance of GCN and R-GCN.}
    \label{tab:detector}
    \begin{tabular}{lllccc}
    \toprule
      ~ & ~ & ~ & Accuracy & Recall & NDCG  \\ 
      \midrule
        Train & Weibo & GCN & 0.5726 & 0.6047 & 0.7570  \\
        ~ & ~ & RGCN & 0.9785 & 0.9698 & 0.7794  \\
        ~ & Twitter & GCN & 0.7253 & 0.5700 & 0.7476  \\
        ~ & ~ & RGCN & 0.8496 & 0.8047 & 0.7623  \\
        ~ & Pheme & GCN & 0.8020 & 0.6855 & 0.5132  \\
        ~ & ~ & RGCN & 0.7912 & 0.5210 & 0.5432  \\
        ~ & Pheme- & GCN & 0.9209 & 0.6703 & 0.4601  \\
        ~ & ~ & RGCN & 0.8211 & 0.9532 & 0.5258  \\
        \midrule
        Test & Weibo & GCN & 0.5506 & 0.5163 & 0.7418  \\ 
        ~ & ~ & RGCN & 0.6647 & 0.5785 & 0.8054  \\ 
        ~ & Twitter & GCN & 0.6268 & 0.4644 & 0.8096  \\
        ~ & ~ & RGCN & 0.6890 & 0.5051 & 0.7152  \\
         ~ & Pheme & GCN & 0.6588 & 0.4949 & 0.4276  \\
        ~ & ~ & RGCN & 0.8496 & 0.8047 & 0.7623  \\
        ~ & Pheme- & GCN & 0.7549 & 0.2521 & 0.3069  \\
        ~ & ~ & RGCN & 0.7053 & 0.3936 & 0.4483  \\
    \bottomrule
    \end{tabular}
\end{table}

\section{Reward Variance Analysis}
\label{app.Variance_analysis}
Given the reward matrix $\boldsymbol{R}=\{r_{e,t}\}\in\mathbb{R}^{E\times T}$, where $r_{e,t}$ indicates the reward in the $t$-th step of the $e$-th trajectory. E is the number of trajectories and $T$ is the length of the horizon.
The variances of the reward with the time-dependent baseline Eq.~(\ref{eq:baseline}) and constant baseline as Eq.~(\ref{eq:const_baseline}) are $ \textnormal{Var}(r_{e,t}-b_t)$ and $ \textnormal{Var}(r_{e,t}-\overline{b})$.
As $\overline{b}$ is a constant, $ \textnormal{Var}(r_{e,t}-\overline{b})= \textnormal{Var}(r_{e,t})$.
According to Theorem~\ref{the.var}, we have $ \textnormal{Var}(r_{e,t})\geq \textnormal{Var}(r_{e,t}-b_t)$. Theorem~\ref{the.var} is proved as follows
\begin{proof}
\begin{align}
    &\sigma^2=\frac{1}{ET}\sum_e\sum_t{
    \left(r_{e,t}-\frac{1}{ET}\sum_e\sum_t{r_{e,t}}\right)}^2, \\
    &{\sigma'}^2=\frac{1}{ET}\sum_e\sum_t{ \left(r_{e,t}-b_t-\frac{1}{ET}\sum_e\sum_t{(r_{e,t}-b_t})\right)}^2,
\end{align}
\begin{equation}
\begin{aligned}
    &\sigma^2-{\sigma'}^2\\
    &=\frac{1}{ET}\sum_e\sum_t{\left(2r_{e,t}-b_t-\frac{1}{ET}\sum_e\sum_t{(2r_{e,t}-b_t})\right)}\\
    &\times\left(b_t-\frac{1}{ET}\sum_e\sum_t{b_t}\right) \\
    &=\frac{1}{ET}\sum_e\sum_t{\left(2r_{e,t}-b_t-\frac{1}{T}\sum_t{b_t}\right)\left(b_t-\frac{1}{T}\sum_t{b_t}\right)}\\
    &=\frac{1}{ET}\sum_e\sum_t{\left(r_{e,t}(2b_t-\frac{2}{T}\sum_t{b_t})\right)}\\
    &+ \frac{1}{T}\sum_t{\left({(\frac{1}{T}\sum_t{b_t})}^2-b_t^2\right)}\\
    &=\frac{1}{T}\sum_t{\left(b_t(2b_t-\frac{2}{T}\sum_t{b_t})+{(\frac{1}{T}\sum_t{b_t})}^2-b_t^2\right)}\\
\end{aligned}
\end{equation}
\begin{equation}
\begin{aligned}
    \notag
    &=\frac{1}{T}\sum_t{\left(b_t-\frac{1}{T}\sum_t{b_t}\right)^2}
    \geq 0,
\end{aligned}
\end{equation}
Thus, $\sigma^2\geq{\sigma'}^2$.
\end{proof}

%% file: appendix_table.tex
\begin{table}
    \centering
    \small
    \caption{\small The designed features for AdRumor-RL on subgraph and node level. \# refers to the number of. $a:b$ means the ratio of $a$ to $b$. * can be replaced with avg/max/min here, which means average/maximum/minimum. The rows related with top-10 important features in Weibo experiments are signed with \colorbox{yellow!40}{highlight}.}
    \label{tab.features}
    \begin{tabular}{l|p{0.63\linewidth}}
    \toprule \midrule
        \textbf{Name} & \textbf{Description} \\ \midrule \midrule
         \multicolumn{2}{c}{\textbf{Subgraph Level}}\\ \midrule
        \multicolumn{2}{c}{\textbf{Structural features}}\\ \midrule
        \rowcolor{yellow!40}n\_nodes & \# nodes.\\
        \rowcolor{yellow!40}n\_edges & \# edges.\\
        clustring\_coefficient & The global clustering coefficient.\\
        *\_degree & The avg/max/min node degrees.  \\\midrule
        \multicolumn{2}{c}{\textbf{Social features}}   \\ \midrule
        message\_ratio & \# message nodes : \# nodes.  \\
        author\_ratio & \# author nodes : \# nodes. 
        \\
        re\_tweeter\_ratio & \# retweeter nodes: \# nodes. 
        \\
        \rowcolor{yellow!40}review\_ratio & The ratio of \# comment nodes.  \\
        bad\_author\_ratio & \# bad author nodes : \# author nodes. 
        \\
        rumor\_ratio & \# rumor nodes : \# message nodes.  \\
        rumor\_retweet\_ratio & \# retweeter nodes who connect to rumor nodes : \# retweeter nodes.  \\
        rumor\_review\_ratio & \# comment nodes who connect to rumor nodes : \# comment nodes.  \\ \midrule
        \multicolumn{2}{c}{\textbf{Influence features}}   \\ \midrule
        *\_author\_inf & \multirow{4}{*}{\shortstack{The avg/max/min influence \\ of author/user/rumor/non-rumor nodes.}}  \\
        *\_user\_inf &   \\ 
        *\_rumor\_inf &   \\ 
        *\_nonrumor\_inf &   \\  
        \midrule
        \multicolumn{2}{c}{\textbf{Attack potential features}}  \\ \midrule
        \rowcolor{yellow!40}*\_target\_suspicious & The avg/max/min probability of target rumors.   \\
        \rowcolor{yellow!40}attack\_degree & \# added edges in the previous steps : horizon $T$.  \\ \midrule 
        \multicolumn{2}{c}{\textbf{Ranking help message features}}  \\ \midrule
        \rowcolor{yellow!40}*\_rhm\_suspicious & The avg/max/min probability of non-target rumors.  \\ \midrule \midrule
        \multicolumn{2}{c}{\textbf{Node Level}}   \\ \midrule
        \multicolumn{2}{c}{\textbf{Structural features}}   \\ \midrule
        degree & The degree of the node.  \\ 
        ego\_n\_edges & \# edges in the ego network.  \\ \midrule
        \multicolumn{2}{c}{\textbf{Social features}}   \\ \midrule
        good\_bad & 0 if the node is good author or non-rumor, 1 if the node is bad author or rumor.  \\ 
        node\_type & The one-hot vector to indicate the node type, including rumor, non-rumor, good author, and bad author.  \\ 
        \rowcolor{yellow!40}ego\_rumor\_ratio & \# rumor nodes : \# nodes in the ego network.  \\ 
        ego\_bu\_ratio & \# bad author nodes : \# nodes in the ego network.  \\ 
        ego\_review\_ratio & \# comment nodes : \# nodes in the ego network.  \\ \midrule
        \multicolumn{2}{c}{\textbf{Influence features}}   \\ \midrule
        node\_inf & The user or message influence of the node.  \\
        ego\_user\_inf & \multirow{2}{*}{\shortstack{The average influence of the user/message \\nodes in the ego network.}}  \\ 
        ego\_message\_inf & \\ \midrule
        \multicolumn{2}{c}{\textbf{Attack potential features}}   \\ \midrule
        \rowcolor{yellow!40}*\_node\_potential & The avg/max/min probability of target rumors within 1-hop insides.   \\
    \midrule
    \end{tabular}
\end{table}
\begin{table}
    \centering
    \small
    \begin{tabular}{l|p{0.63\linewidth}}
        \midrule
        \rowcolor{yellow!40}*\_neighbor\_suspicious & The avg/max/min probability of target rumors within the node $k$-hop insides.  \\ 
        \rowcolor{yellow!40}*\_node\_attack\_degree & \# added edges that connect to the node in the previous steps : horizon $T$.  \\ 
        n\_targets & \# target rumors within the node $k$-hop insides.  \\ 
        n\_targets\_distance & The average distance from the node to the target rumors within the node n-hop insides.  \\ \midrule
        \multicolumn{2}{c}{\textbf{Ranking help message features}}   \\ \midrule
        \rowcolor{yellow!40} *\_rhm\_suspicious & The avg/max/min probability of non-target rumors within the node n-hop insides.  \\ 
    \bottomrule
    \end{tabular}
\end{table}

%% file: sample-sigconf.bbl

\begin{thebibliography}{41}


\ifx \showCODEN    \undefined \def \showCODEN     #1{\unskip}     \fi
\ifx \showDOI      \undefined \def \showDOI       #1{#1}\fi
\ifx \showISBNx    \undefined \def \showISBNx     #1{\unskip}     \fi
\ifx \showISBNxiii \undefined \def \showISBNxiii  #1{\unskip}     \fi
\ifx \showISSN     \undefined \def \showISSN      #1{\unskip}     \fi
\ifx \showLCCN     \undefined \def \showLCCN      #1{\unskip}     \fi
\ifx \shownote     \undefined \def \shownote      #1{#1}          \fi
\ifx \showarticletitle \undefined \def \showarticletitle #1{#1}   \fi
\ifx \showURL      \undefined \def \showURL       {\relax}        \fi
\providecommand\bibfield[2]{#2}
\providecommand\bibinfo[2]{#2}
\providecommand\natexlab[1]{#1}
\providecommand\showeprint[2][]{arXiv:#2}

\bibitem[Alharin et~al\mbox{.}(2020)]%
        {Alharin2020}
\bibfield{author}{\bibinfo{person}{Alnour Alharin}, \bibinfo{person}{Thanh-Nam
  Doan}, {and} \bibinfo{person}{Mina Sartipi}.}
  \bibinfo{year}{2020}\natexlab{}.
\newblock \showarticletitle{{Reinforcement Learning Interpretation Methods: A
  Survey}}.
\newblock \bibinfo{journal}{\emph{IEEE Access}}  \bibinfo{volume}{8}
  (\bibinfo{year}{2020}), \bibinfo{pages}{171058--171077}.
\newblock


\bibitem[Bakshy et~al\mbox{.}(2011)]%
        {Bakshy2011}
\bibfield{author}{\bibinfo{person}{Eytan Bakshy}, \bibinfo{person}{Jake~M.
  Hofman}, \bibinfo{person}{Winter~A. Mason}, {and} \bibinfo{person}{Duncan~J.
  Watts}.} \bibinfo{year}{2011}\natexlab{}.
\newblock \showarticletitle{Everyone's an Influencer: Quantifying Influence on
  Twitter}. In \bibinfo{booktitle}{\emph{WSDM '11}}. \bibinfo{pages}{65–74}.
\newblock


\bibitem[Bian et~al\mbox{.}(2020)]%
        {3_Bian_Xiao_Xu_Zhao_Huang_Rong_Huang_2020}
\bibfield{author}{\bibinfo{person}{Tian Bian}, \bibinfo{person}{Xi Xiao},
  \bibinfo{person}{Tingyang Xu}, \bibinfo{person}{Peilin Zhao},
  \bibinfo{person}{Wenbing Huang}, \bibinfo{person}{Yu Rong}, {and}
  \bibinfo{person}{Junzhou Huang}.} \bibinfo{year}{2020}\natexlab{}.
\newblock \showarticletitle{Rumor Detection on Social Media with Bi-Directional
  Graph Convolutional Networks}.
\newblock \bibinfo{journal}{\emph{AAAI}} \bibinfo{volume}{34},
  \bibinfo{number}{01} (\bibinfo{year}{2020}), \bibinfo{pages}{549--556}.
\newblock


\bibitem[Bishop and Nasrabadi(2006)]%
        {prml}
\bibfield{author}{\bibinfo{person}{Christopher~M Bishop} {and}
  \bibinfo{person}{Nasser~M Nasrabadi}.} \bibinfo{year}{2006}\natexlab{}.
\newblock \bibinfo{booktitle}{\emph{Pattern recognition and machine learning}}.
  Vol.~\bibinfo{volume}{4}.
\newblock \bibinfo{publisher}{Springer}.
\newblock


\bibitem[Bojchevski and G{\"u}nnemann(2019)]%
        {11_pmlr-v97-bojchevski19a}
\bibfield{author}{\bibinfo{person}{Aleksandar Bojchevski} {and}
  \bibinfo{person}{Stephan G{\"u}nnemann}.} \bibinfo{year}{2019}\natexlab{}.
\newblock \showarticletitle{Adversarial Attacks on Node Embeddings via Graph
  Poisoning}. In \bibinfo{booktitle}{\emph{PMLR 2019}},
  Vol.~\bibinfo{volume}{97}. \bibinfo{pages}{695--704}.
\newblock


\bibitem[Dai et~al\mbox{.}(2018)]%
        {6_pmlr-v80-dai18b}
\bibfield{author}{\bibinfo{person}{Hanjun Dai}, \bibinfo{person}{Hui Li},
  \bibinfo{person}{Tian Tian}, \bibinfo{person}{Xin Huang},
  \bibinfo{person}{Lin Wang}, \bibinfo{person}{Jun Zhu}, {and}
  \bibinfo{person}{Le Song}.} \bibinfo{year}{2018}\natexlab{}.
\newblock \showarticletitle{Adversarial Attack on Graph Structured Data},
  Vol.~\bibinfo{volume}{80}. \bibinfo{publisher}{PMLR 2018},
  \bibinfo{pages}{1115--1124}.
\newblock


\bibitem[Foerster et~al\mbox{.}(2018)]%
        {credit2}
\bibfield{author}{\bibinfo{person}{Jakob~N. Foerster}, \bibinfo{person}{Gregory
  Farquhar}, \bibinfo{person}{Triantafyllos Afouras}, \bibinfo{person}{Nantas
  Nardelli}, {and} \bibinfo{person}{Shimon Whiteson}.}
  \bibinfo{year}{2018}\natexlab{}.
\newblock \showarticletitle{Counterfactual Multi-Agent Policy Gradients}. In
  \bibinfo{booktitle}{\emph{Proceedings of the Thirty-Second {AAAI} Conference
  on Artificial Intelligence,(AAAI-18)}},
  \bibfield{editor}{\bibinfo{person}{Sheila~A. McIlraith} {and}
  \bibinfo{person}{Kilian~Q. Weinberger}} (Eds.). \bibinfo{publisher}{{AAAI}
  Press}, \bibinfo{pages}{2974--2982}.
\newblock


\bibitem[Garg et~al\mbox{.}(2020)]%
        {Garg2020}
\bibfield{author}{\bibinfo{person}{Vikas~K Garg}, \bibinfo{person}{Stefanie
  Jegelka}, {and} \bibinfo{person}{Tommi~S Jaakkola}.}
  \bibinfo{year}{2020}\natexlab{}.
\newblock \showarticletitle{{Generalization and Representational Limits of
  Graph Neural Networks}}.
\newblock \bibinfo{journal}{\emph{CoRR}}  \bibinfo{volume}{abs/2002.0}
  (\bibinfo{year}{2020}).
\newblock


\bibitem[Greensmith et~al\mbox{.}(2004)]%
        {Greensmith2004VarianceRT}
\bibfield{author}{\bibinfo{person}{Evan Greensmith}, \bibinfo{person}{Peter~L.
  Bartlett}, {and} \bibinfo{person}{Jonathan Baxter}.}
  \bibinfo{year}{2004}\natexlab{}.
\newblock \showarticletitle{Variance Reduction Techniques for Gradient
  Estimates in Reinforcement Learning}. In \bibinfo{booktitle}{\emph{J. Mach.
  Learn. Res.}}
\newblock


\bibitem[Hodas and Lerman(2012)]%
        {Hodas2012}
\bibfield{author}{\bibinfo{person}{Nathan Hodas} {and}
  \bibinfo{person}{Kristina Lerman}.} \bibinfo{year}{2012}\natexlab{}.
\newblock \showarticletitle{How Visibility and Divided Attention Constrain
  Social Contagion}.
\newblock  (\bibinfo{date}{05} \bibinfo{year}{2012}).
\newblock


\bibitem[Kimura and Kobayashi(1998)]%
        {kimura1998reinforcement}
\bibfield{author}{\bibinfo{person}{Hajime Kimura} {and}
  \bibinfo{person}{Shigenobu Kobayashi}.} \bibinfo{year}{1998}\natexlab{}.
\newblock \showarticletitle{Reinforcement learning for continuous action using
  stochastic gradient ascent}.
\newblock \bibinfo{journal}{\emph{Intelligent Autonomous Systems (IAS-5)}}
  (\bibinfo{year}{1998}), \bibinfo{pages}{288--295}.
\newblock


\bibitem[Kipf and Welling(2017)]%
        {12_DBLP:conf/iclr/KipfW17}
\bibfield{author}{\bibinfo{person}{Thomas~N. Kipf} {and} \bibinfo{person}{Max
  Welling}.} \bibinfo{year}{2017}\natexlab{}.
\newblock \showarticletitle{Semi-Supervised Classification with Graph
  Convolutional Networks}. In \bibinfo{booktitle}{\emph{ICLR 2017}}.
\newblock


\bibitem[Konda and Tsitsiklis(2000)]%
        {konda2000actor}
\bibfield{author}{\bibinfo{person}{Vijay~R Konda} {and} \bibinfo{person}{John~N
  Tsitsiklis}.} \bibinfo{year}{2000}\natexlab{}.
\newblock \showarticletitle{Actor-critic algorithms}. In
  \bibinfo{booktitle}{\emph{Advances in neural information processing
  systems}}. \bibinfo{pages}{1008--1014}.
\newblock


\bibitem[Lee et~al\mbox{.}(2019)]%
        {DBLP:conf/cikm/LeeRKKKR19}
\bibfield{author}{\bibinfo{person}{John~Boaz Lee}, \bibinfo{person}{Ryan~A.
  Rossi}, \bibinfo{person}{Xiangnan Kong}, \bibinfo{person}{Sungchul Kim},
  \bibinfo{person}{Eunyee Koh}, {and} \bibinfo{person}{Anup Rao}.}
  \bibinfo{year}{2019}\natexlab{}.
\newblock \showarticletitle{Graph Convolutional Networks with Motif-based
  Attention}. In \bibinfo{booktitle}{\emph{Proceedings of the 28th {ACM}
  International Conference on Information and Knowledge Management, {CIKM}
  2019, Beijing, China, November 3-7, 2019}},
  \bibfield{editor}{\bibinfo{person}{Wenwu Zhu}, \bibinfo{person}{Dacheng Tao},
  \bibinfo{person}{Xueqi Cheng}, \bibinfo{person}{Peng Cui},
  \bibinfo{person}{Elke~A. Rundensteiner}, \bibinfo{person}{David Carmel},
  \bibinfo{person}{Qi~He}, {and} \bibinfo{person}{Jeffrey~Xu Yu}} (Eds.).
  \bibinfo{publisher}{{ACM}}, \bibinfo{pages}{499--508}.
\newblock


\bibitem[Li et~al\mbox{.}(2010)]%
        {15_10.1145/1772690.1772758}
\bibfield{author}{\bibinfo{person}{Lihong Li}, \bibinfo{person}{Wei Chu},
  \bibinfo{person}{John Langford}, {and} \bibinfo{person}{Robert~E. Schapire}.}
  \bibinfo{year}{2010}\natexlab{}.
\newblock \showarticletitle{A Contextual-Bandit Approach to Personalized News
  Article Recommendation}. \bibinfo{publisher}{WWW '10},
  \bibinfo{pages}{661–670}.
\newblock


\bibitem[Lu and Li(2020)]%
        {4_lu-li-2020-gcan}
\bibfield{author}{\bibinfo{person}{Yi-Ju Lu} {and} \bibinfo{person}{Cheng-Te
  Li}.} \bibinfo{year}{2020}\natexlab{}.
\newblock \showarticletitle{{GCAN}: Graph-aware Co-Attention Networks for
  Explainable Fake News Detection on Social Media}. In
  \bibinfo{booktitle}{\emph{ACL 2020}}. \bibinfo{pages}{505--514}.
\newblock


\bibitem[Ma et~al\mbox{.}(2020)]%
        {7_NEURIPS2020_32bb90e8}
\bibfield{author}{\bibinfo{person}{Jiaqi Ma}, \bibinfo{person}{Shuangrui Ding},
  {and} \bibinfo{person}{Qiaozhu Mei}.} \bibinfo{year}{2020}\natexlab{}.
\newblock \showarticletitle{Towards More Practical Adversarial Attacks on Graph
  Neural Networks}. In \bibinfo{booktitle}{\emph{Advances in Neural Information
  Processing Systems}}, \bibfield{editor}{\bibinfo{person}{H.~Larochelle},
  \bibinfo{person}{M.~Ranzato}, \bibinfo{person}{R.~Hadsell},
  \bibinfo{person}{M.~F. Balcan}, {and} \bibinfo{person}{H.~Lin}} (Eds.),
  Vol.~\bibinfo{volume}{33}. \bibinfo{pages}{4756--4766}.
\newblock


\bibitem[Ma et~al\mbox{.}(2017)]%
        {TwitterDataset}
\bibfield{author}{\bibinfo{person}{Jing Ma}, \bibinfo{person}{Wei Gao}, {and}
  \bibinfo{person}{Kam{-}Fai Wong}.} \bibinfo{year}{2017}\natexlab{}.
\newblock \showarticletitle{Detect Rumors in Microblog Posts Using Propagation
  Structure via Kernel Learning}. In \bibinfo{booktitle}{\emph{Proceedings of
  the 55th Annual Meeting of the Association for Computational Linguistics,
  {ACL} 2017, Vancouver, Canada, July 30 - August 4, Volume 1: Long Papers}},
  \bibfield{editor}{\bibinfo{person}{Regina Barzilay} {and}
  \bibinfo{person}{Min{-}Yen Kan}} (Eds.). \bibinfo{pages}{708--717}.
\newblock


\bibitem[Mao et~al\mbox{.}(2019)]%
        {Mao2019VarianceRF}
\bibfield{author}{\bibinfo{person}{Hongzi Mao},
  \bibinfo{person}{Shaileshh~Bojja Venkatakrishnan}, \bibinfo{person}{Malte
  Schwarzkopf}, {and} \bibinfo{person}{Mohammad Alizadeh}.}
  \bibinfo{year}{2019}\natexlab{}.
\newblock \showarticletitle{{Variance Reduction for Reinforcement Learning in
  Input-Driven Environments}}.
\newblock \bibinfo{journal}{\emph{ICLR}} (\bibinfo{year}{2019}).
\newblock


\bibitem[Marbach and Tsitsiklis(2001)]%
        {marbach2001simulation}
\bibfield{author}{\bibinfo{person}{Peter Marbach} {and} \bibinfo{person}{John~N
  Tsitsiklis}.} \bibinfo{year}{2001}\natexlab{}.
\newblock \showarticletitle{Simulation-based optimization of Markov reward
  processes}.
\newblock \bibinfo{journal}{\emph{IEEE Trans. Automat. Control}}
  \bibinfo{volume}{46}, \bibinfo{number}{2} (\bibinfo{year}{2001}),
  \bibinfo{pages}{191--209}.
\newblock


\bibitem[Mnih et~al\mbox{.}(2015)]%
        {Mnih2015}
\bibfield{author}{\bibinfo{person}{Volodymyr Mnih}, \bibinfo{person}{Koray
  Kavukcuoglu}, \bibinfo{person}{David Silver}, \bibinfo{person}{Andrei~A
  Rusu}, \bibinfo{person}{Joel Veness}, \bibinfo{person}{Marc~G Bellemare},
  \bibinfo{person}{Alex Graves}, \bibinfo{person}{Martin Riedmiller},
  \bibinfo{person}{Andreas~K Fidjeland}, \bibinfo{person}{Georg Ostrovski},
  \bibinfo{person}{Stig Petersen}, \bibinfo{person}{Charles Beattie},
  \bibinfo{person}{Amir Sadik}, \bibinfo{person}{Ioannis Antonoglou},
  \bibinfo{person}{Helen King}, \bibinfo{person}{Dharshan Kumaran},
  \bibinfo{person}{Daan Wierstra}, \bibinfo{person}{Shane Legg}, {and}
  \bibinfo{person}{Demis Hassabis}.} \bibinfo{year}{2015}\natexlab{}.
\newblock \showarticletitle{{Human-level control through deep reinforcement
  learning}}.
\newblock \bibinfo{journal}{\emph{Nature}} \bibinfo{volume}{518},
  \bibinfo{number}{7540} (\bibinfo{date}{feb} \bibinfo{year}{2015}),
  \bibinfo{pages}{529--533}.
\newblock


\bibitem[Oono and Suzuki(2020)]%
        {DBLP:conf/iclr/OonoS20}
\bibfield{author}{\bibinfo{person}{Kenta Oono} {and} \bibinfo{person}{Taiji
  Suzuki}.} \bibinfo{year}{2020}\natexlab{}.
\newblock \showarticletitle{Graph Neural Networks Exponentially Lose Expressive
  Power for Node Classification}. In \bibinfo{booktitle}{\emph{8th
  International Conference on Learning Representations, {ICLR} 2020, Addis
  Ababa, Ethiopia, April 26-30, 2020}}. \bibinfo{publisher}{OpenReview.net}.
\newblock


\bibitem[Qazaz et~al\mbox{.}(1997)]%
        {Qazaz1997}
\bibfield{author}{\bibinfo{person}{Cazhaow~S. Qazaz},
  \bibinfo{person}{Christopher K.~I. Williams}, {and}
  \bibinfo{person}{Christopher~M. Bishop}.} \bibinfo{year}{1997}\natexlab{}.
\newblock \bibinfo{booktitle}{\emph{An Upper Bound on the Bayesian Error Bars
  for Generalized Linear Regression}}.
\newblock \bibinfo{publisher}{Springer US}, \bibinfo{address}{Boston, MA},
  \bibinfo{pages}{295--299}.
\newblock
\showISBNx{978-1-4615-6099-9}


\bibitem[Ribeiro et~al\mbox{.}(2016)]%
        {Ribeiro2016}
\bibfield{author}{\bibinfo{person}{Marco~Tulio Ribeiro},
  \bibinfo{person}{Sameer Singh}, {and} \bibinfo{person}{Carlos Guestrin}.}
  \bibinfo{year}{2016}\natexlab{}.
\newblock \showarticletitle{{"Why Should I Trust You?": Explaining the
  Predictions of Any Classifier}} \emph{(\bibinfo{series}{KDD})}.
  \bibinfo{publisher}{ACM}, \bibinfo{pages}{1135--1144}.
\newblock


\bibitem[Ruiz et~al\mbox{.}(2021)]%
        {Ruiz2021GraphNN}
\bibfield{author}{\bibinfo{person}{Luana Ruiz}, \bibinfo{person}{Fernando
  Gama}, {and} \bibinfo{person}{Alejandro Ribeiro}.}
  \bibinfo{year}{2021}\natexlab{}.
\newblock \showarticletitle{Graph Neural Networks: Architectures, Stability,
  and Transferability}.
\newblock \bibinfo{journal}{\emph{Proc. IEEE}}  \bibinfo{volume}{109}
  (\bibinfo{year}{2021}), \bibinfo{pages}{660--682}.
\newblock


\bibitem[Schlichtkrull et~al\mbox{.}(2018)]%
        {14_10.1007/978-3-319-93417-4_38}
\bibfield{author}{\bibinfo{person}{Michael Schlichtkrull},
  \bibinfo{person}{Thomas~N. Kipf}, \bibinfo{person}{Peter Bloem},
  \bibinfo{person}{Rianne van den Berg}, \bibinfo{person}{Ivan Titov}, {and}
  \bibinfo{person}{Max Welling}.} \bibinfo{year}{2018}\natexlab{}.
\newblock \showarticletitle{Modeling Relational Data with Graph Convolutional
  Networks}. In \bibinfo{booktitle}{\emph{The Semantic Web}},
  \bibfield{editor}{\bibinfo{person}{Aldo Gangemi}, \bibinfo{person}{Roberto
  Navigli}, \bibinfo{person}{Maria-Esther Vidal}, \bibinfo{person}{Pascal
  Hitzler}, \bibinfo{person}{Rapha{\"e}l Troncy}, \bibinfo{person}{Laura
  Hollink}, \bibinfo{person}{Anna Tordai}, {and} \bibinfo{person}{Mehwish
  Alam}} (Eds.). \bibinfo{pages}{593--607}.
\newblock


\bibitem[Seo et~al\mbox{.}(2019)]%
        {credit1}
\bibfield{author}{\bibinfo{person}{Minah Seo}, \bibinfo{person}{Luiz~Felipe
  Vecchietti}, \bibinfo{person}{Sangkeum Lee}, {and} \bibinfo{person}{Dongsoo
  Har}.} \bibinfo{year}{2019}\natexlab{}.
\newblock \showarticletitle{Rewards Prediction-Based Credit Assignment for
  Reinforcement Learning With Sparse Binary Rewards}.
\newblock \bibinfo{journal}{\emph{IEEE Access}}  \bibinfo{volume}{7}
  (\bibinfo{year}{2019}), \bibinfo{pages}{118776--118791}.
\newblock


\bibitem[Song et~al\mbox{.}(2021)]%
        {weibo}
\bibfield{author}{\bibinfo{person}{Changhe Song}, \bibinfo{person}{Cheng Yang},
  \bibinfo{person}{Huimin Chen}, \bibinfo{person}{Cunchao Tu},
  \bibinfo{person}{Zhiyuan Liu}, {and} \bibinfo{person}{Maosong Sun}.}
  \bibinfo{year}{2021}\natexlab{}.
\newblock \showarticletitle{CED: Credible Early Detection of Social Media
  Rumors}.
\newblock \bibinfo{journal}{\emph{TKDE 2021}} \bibinfo{volume}{33},
  \bibinfo{number}{8} (\bibinfo{year}{2021}), \bibinfo{pages}{3035--3047}.
\newblock


\bibitem[Sutton and Barto(2018)]%
        {Sutton1998}
\bibfield{author}{\bibinfo{person}{Richard~S Sutton} {and}
  \bibinfo{person}{Andrew~G Barto}.} \bibinfo{year}{2018}\natexlab{}.
\newblock \bibinfo{booktitle}{\emph{Reinforcement learning: An introduction}}.
\newblock \bibinfo{publisher}{MIT press}.
\newblock


\bibitem[Sutton et~al\mbox{.}(2000)]%
        {sutton2000policy}
\bibfield{author}{\bibinfo{person}{Richard~S Sutton}, \bibinfo{person}{David~A
  McAllester}, \bibinfo{person}{Satinder~P Singh}, {and}
  \bibinfo{person}{Yishay Mansour}.} \bibinfo{year}{2000}\natexlab{}.
\newblock \showarticletitle{Policy gradient methods for reinforcement learning
  with function approximation}. In \bibinfo{booktitle}{\emph{Advances in neural
  information processing systems}}. \bibinfo{pages}{1057--1063}.
\newblock


\bibitem[Tucker et~al\mbox{.}(2018)]%
        {Tucker2018TheMO}
\bibfield{author}{\bibinfo{person}{G. Tucker}, \bibinfo{person}{Surya
  Bhupatiraju}, \bibinfo{person}{Shixiang~Shane Gu},
  \bibinfo{person}{Richard~E. Turner}, \bibinfo{person}{Zoubin Ghahramani},
  {and} \bibinfo{person}{Sergey Levine}.} \bibinfo{year}{2018}\natexlab{}.
\newblock \showarticletitle{The Mirage of Action-Dependent Baselines in
  Reinforcement Learning}.
\newblock \bibinfo{journal}{\emph{ArXiv}}  \bibinfo{volume}{abs/1802.10031}
  (\bibinfo{year}{2018}).
\newblock


\bibitem[Wang et~al\mbox{.}(2020)]%
        {wang2020long}
\bibfield{author}{\bibinfo{person}{Ruosong Wang}, \bibinfo{person}{Simon~S.
  Du}, \bibinfo{person}{Lin~F. Yang}, {and} \bibinfo{person}{Sham~M. Kakade}.}
  \bibinfo{year}{2020}\natexlab{}.
\newblock \bibinfo{title}{Is Long Horizon Reinforcement Learning More Difficult
  Than Short Horizon Reinforcement Learning?}
\newblock
\newblock
\showeprint[arxiv]{2005.00527}~[cs.LG]


\bibitem[Weaver and Tao(2001)]%
        {Weaver2001TheOR}
\bibfield{author}{\bibinfo{person}{Lex Weaver} {and} \bibinfo{person}{Nigel
  Tao}.} \bibinfo{year}{2001}\natexlab{}.
\newblock \showarticletitle{The Optimal Reward Baseline for Gradient-Based
  Reinforcement Learning}. In \bibinfo{booktitle}{\emph{UAI}}.
\newblock


\bibitem[Williams(1992)]%
        {williams1992simple}
\bibfield{author}{\bibinfo{person}{Ronald~J Williams}.}
  \bibinfo{year}{1992}\natexlab{}.
\newblock \showarticletitle{Simple statistical gradient-following algorithms
  for connectionist reinforcement learning}.
\newblock \bibinfo{journal}{\emph{Machine learning}} \bibinfo{volume}{8},
  \bibinfo{number}{3} (\bibinfo{year}{1992}), \bibinfo{pages}{229--256}.
\newblock


\bibitem[Wu et~al\mbox{.}(2018)]%
        {Wu2018VarianceRF}
\bibfield{author}{\bibinfo{person}{Cathy Wu}, \bibinfo{person}{Aravind
  Rajeswaran}, \bibinfo{person}{Yan Duan}, \bibinfo{person}{Vikash Kumar},
  \bibinfo{person}{Alexandre~M. Bayen}, \bibinfo{person}{Sham~M. Kakade},
  \bibinfo{person}{Igor Mordatch}, {and} \bibinfo{person}{P. Abbeel}.}
  \bibinfo{year}{2018}\natexlab{}.
\newblock \showarticletitle{Variance Reduction for Policy Gradient with
  Action-Dependent Factorized Baselines}.
\newblock \bibinfo{journal}{\emph{ArXiv}}  \bibinfo{volume}{abs/1803.07246}
  (\bibinfo{year}{2018}).
\newblock


\bibitem[Wu et~al\mbox{.}(2019)]%
        {10_ijcai2019-669}
\bibfield{author}{\bibinfo{person}{Huijun Wu}, \bibinfo{person}{Chen Wang},
  \bibinfo{person}{Yuriy Tyshetskiy}, \bibinfo{person}{Andrew Docherty},
  \bibinfo{person}{Kai Lu}, {and} \bibinfo{person}{Liming Zhu}.}
  \bibinfo{year}{2019}\natexlab{}.
\newblock \showarticletitle{Adversarial Examples for Graph Data: Deep Insights
  into Attack and Defense}. In \bibinfo{booktitle}{\emph{IJCAI-19}}.
  \bibinfo{pages}{4816--4823}.
\newblock


\bibitem[Xu et~al\mbox{.}(2019)]%
        {xu2018how}
\bibfield{author}{\bibinfo{person}{Keyulu Xu}, \bibinfo{person}{Weihua Hu},
  \bibinfo{person}{Jure Leskovec}, {and} \bibinfo{person}{Stefanie Jegelka}.}
  \bibinfo{year}{2019}\natexlab{}.
\newblock \showarticletitle{How Powerful are Graph Neural Networks?}. In
  \bibinfo{booktitle}{\emph{International Conference on Learning
  Representations}}.
\newblock
\urldef\tempurl%
\url{https://openreview.net/forum?id=ryGs6iA5Km}
\showURL{%
\tempurl}


\bibitem[Yang et~al\mbox{.}(2020)]%
        {1_ijcai2020-197}
\bibfield{author}{\bibinfo{person}{Xiaoyu Yang}, \bibinfo{person}{Yuefei Lyu},
  \bibinfo{person}{Tian Tian}, \bibinfo{person}{Yifei Liu},
  \bibinfo{person}{Yudong Liu}, {and} \bibinfo{person}{Xi Zhang}.}
  \bibinfo{year}{2020}\natexlab{}.
\newblock \showarticletitle{Rumor Detection on Social Media with Graph
  Structured Adversarial Learning}. In \bibinfo{booktitle}{\emph{IJCAI-20}}.
  \bibinfo{pages}{1417--1423}.
\newblock


\bibitem[Zubiaga et~al\mbox{.}(2016)]%
        {pheme}
\bibfield{author}{\bibinfo{person}{Arkaitz Zubiaga}, \bibinfo{person}{Maria
  Liakata}, {and} \bibinfo{person}{Rob Procter}.}
  \bibinfo{year}{2016}\natexlab{}.
\newblock \showarticletitle{Learning Reporting Dynamics during Breaking News
  for Rumour Detection in Social Media}.
\newblock \bibinfo{journal}{\emph{CoRR}}  \bibinfo{volume}{abs/1610.07363}
  (\bibinfo{year}{2016}).
\newblock
\showeprint[arXiv]{1610.07363}
\urldef\tempurl%
\url{http://arxiv.org/abs/1610.07363}
\showURL{%
\tempurl}


\bibitem[Z\"{u}gner et~al\mbox{.}(2018)]%
        {8_10.1145/3219819.3220078}
\bibfield{author}{\bibinfo{person}{Daniel Z\"{u}gner}, \bibinfo{person}{Amir
  Akbarnejad}, {and} \bibinfo{person}{Stephan G\"{u}nnemann}.}
  \bibinfo{year}{2018}\natexlab{}.
\newblock \showarticletitle{Adversarial Attacks on Neural Networks for Graph
  Data}. In \bibinfo{booktitle}{\emph{KDD '18}}. \bibinfo{pages}{2847–2856}.
\newblock


\bibitem[Z{\"{u}}gner and G{\"{u}}nnemann(2019)]%
        {9_12_DBLP:conf/iclr/ZugnerG19}
\bibfield{author}{\bibinfo{person}{Daniel Z{\"{u}}gner} {and}
  \bibinfo{person}{Stephan G{\"{u}}nnemann}.} \bibinfo{year}{2019}\natexlab{}.
\newblock \showarticletitle{Adversarial Attacks on Graph Neural Networks via
  Meta Learning}. In \bibinfo{booktitle}{\emph{ICLR 2019}}.
\newblock


\end{thebibliography}
